\documentclass[11pt]{article}
\UseRawInputEncoding

\usepackage[preprint]{acl}
\usepackage{amsmath}
\usepackage{times}
\usepackage{titletoc}
\usepackage{xurl}
\usepackage{hyperref}
\usepackage{latexsym}
\usepackage[normalem]{ulem}
\usepackage{multirow}
\usepackage{array}
\usepackage{graphicx}
\usepackage{subfigure}  
\usepackage{booktabs}
\usepackage{xspace}
\usepackage{colortbl}  
\usepackage{xcolor}    
\usepackage{tikz}
\usepackage{fp}
\usepackage{amssymb}
\usepackage[shortlabels]{enumitem}
\usepackage{tcolorbox}
\tcbuselibrary{breakable}

\usepackage[table]{xcolor}
\usepackage{pgfplotstable}
\newtcolorbox{responsebox}[1]{
    breakable,
    colback=gray!5,
    colframe=gray!50,
    title=#1,
    boxrule=0.5pt,
    arc=2pt,
    left=6pt,
    right=6pt,
    top=6pt,
    bottom=6pt
  }

\usepackage{soul}
\usepackage{fvextra}
\setuldepth{La}
\sethlcolor{yellow}

\newcommand{\toolname}{\textsc{RESCUE-Bench}\xspace}

\pgfplotstableset{
    color cells/.style={
        col sep=comma,
        string type,
        postproc cell content/.code={%
                \pgfkeysalso{@cell content=\rule{0cm}{2.4ex}\cellcolor{blue!##1}\pgfmathtruncatemacro\number{##1}\ifnum\number>50\color{white}\fi##1}%
                },
        columns/x/.style={
            column name={},
            postproc cell content/.code={}
        }
    }
}

\usepackage[T1]{fontenc}

\usepackage[utf8]{inputenc}

\usepackage{microtype}

\usepackage{inconsolata}

\usepackage{graphicx}

\title{\toolname{}: Towards Relation-Aware Multi-Party Emotional Support Conversation Systems}

\author{
  \textbf{Haichuan Hu\textsuperscript{1}},
  \textbf{Yang Xiao\textsuperscript{1}},
  \textbf{Mingni Tang\textsuperscript{1}},
  \textbf{Jiawen Duan\textsuperscript{1}}, 
  \textbf{Quanjun Zhang\textsuperscript{5}}\\
  \textbf{Congqing He\textsuperscript{1}},
  \textbf{Hao Zhang\textsuperscript{6}}, 
  \textbf{Jiashuo Wang\textsuperscript{1}},
  \textbf{Johan F. Hoorn\textsuperscript{1,2,3,4}},
  \textbf{Wenjie Li\textsuperscript{1}} \\
  \textsuperscript{1}Department of Computing, Hong Kong Polytechnic University \\
  \textsuperscript{2}School of Design, Hong Kong Polytechnic University \\
  \textsuperscript{3}Research Institute for Quantum Technology, Hong Kong Polytechnic University \\
  \textsuperscript{4}Department of Communication Science, Vrije Universiteit Amsterdam \\
  \textsuperscript{5}School of Computer Science and Engineering, Nanjing University of Science and Technology \\
  \textsuperscript{6}XU Exponential University of Applied Sciences
}

\begin{document}
\maketitle
\begin{abstract}
Existing emotional support conversation systems mainly focus on one-on-one seeker-supporter interactions and individual emotional states, leaving interpersonal relations in multi-party scenarios underexplored. In this work, we introduce \textit{relation-aware emotional support conversation}, a new task that evaluates whether LLMs can capture and utilize the evolving dynamics of relationships to offer more effective emotional support. We construct \toolname{} 
(\underline{R}elation-aware \underline{E}motional \underline{S}upport 
\underline{C}onversation \underline{U}nderstanding and 
\underline{E}valuation Benchmark) from real couple and family interview conversations, 
containing 191 samples, 7,079 annotated turns, and 1,064.8 minutes of video. Based on rich annotations of socio-emotional and support-related dynamics, \toolname{} defines six tasks that evaluate two core capabilities required for relation-aware emotional support: \textit{Relational Understanding} and \textit{Relation-Sensitive Support}. Experiments with ten LLMs show that current models perform relatively well on tasks relying on local emotional or intervention cues, but struggle with relation-intensive tasks such as relation pattern prediction, viewpoint prediction, and support strategy prediction. These findings reveal the limitations of current LLMs in modeling interpersonal relations and making relation-sensitive support decisions. We release our code and data on Github\footnote{\url{https://github.com/Tomsawyerhu/RESCUE-bench}} and Huggingface\footnote{\url{https://huggingface.co/datasets/tomhu/relation_therapy}}.
\end{abstract} 

\section{Introduction}
Large Language Model (LLM)-based Emotional Support Conversation (ESC) systems have made significant progress in recent years. By leveraging LLMs as emotional supporters, ESC systems can better understand users' emotional needs and personality traits, and provide high-quality empathetic responses.

\begin{figure*}[htbp]
  \centering
  \includegraphics[width=1.0\linewidth]{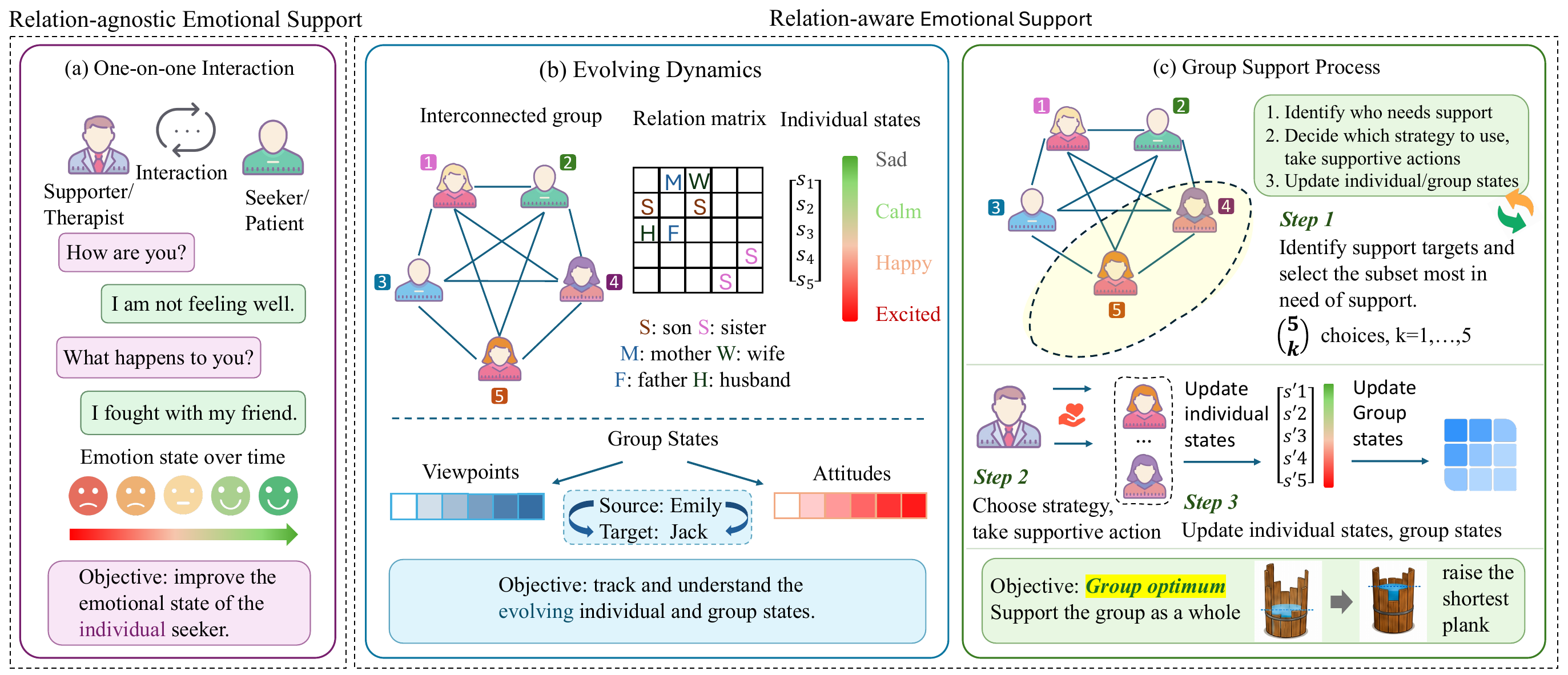}
  \caption{
  Comparison between relation-agnostic ESC (a) and relation-aware ESC (b, c). Unlike relation-agnostic settings, which focus on improving individual emotional states, relation-aware ESC aims to improve the collective support outcome by prioritizing emotionally vulnerable members and adapting support strategies based on interpersonal relationships, relational tensions, and evolving individual/interpersonal dynamics.
  }
  \label{fig:task_def}
\end{figure*}

Existing ESC research~\citep{madani2025steering,xu2025multiagentesc,ye2025sweetiechat} mainly focuses on one-on-one seeker-provider interactions (Figure~\ref{fig:task_def}, left), as exemplified by ESConv~\citep{liu2021towards}. Beyond this setting, multi-party support scenarios~\citep{shalaby2020peer,marshall2024understanding,yuan2025online,prescott2017peer,tracy2016benefits} are often conceptualized as parallel extensions of single-person ESC, in which multiple participants receive support independently without explicitly modeling their interpersonal relationships. In such relation-agnostic support settings, the supporter primarily focuses on individual emotional states, without explicitly accounting for the relationships among seekers or the effects of the evolving relational dynamics on the overall support process.


In contrast, relation-aware ESC (Figure~\ref{fig:task_def}, right) differs from relation-agnostic settings in both its \textbf{objective} and \textbf{support process}. In terms of the objective, relation-agnostic ESC focuses on improving individual seeker's emotional state, whereas relation-aware ESC aims to provide support that benefits the group as a whole, by addressing vulnerable members’ distress while accounting for interpersonal tensions and dependencies. This intuition echoes both the \textit{barrel effect}~\citep{van1999origin,tang2021finding} and the \textit{ripple effect}~\citep{barsade2002ripple}: group-level support may be constrained by vulnerable members' unresolved distress and by emotions or tensions that spread through key interpersonal relations~\citep{barsade2002ripple,felps2006and}, rather than by average individual improvement alone. In terms of the support process, relation-agnostic ESC mainly considers the direct effect of a support action on the target seeker. By contrast, multi-party relation-aware ESC must further account for its indirect influence on other seekers and their interpersonal relations~\citep{reeck2016social,barthel2018interpersonal}.

\begin{table*}[t]
\centering
\scriptsize
\setlength{\tabcolsep}{2.2pt}
\renewcommand{\arraystretch}{1.12}
\caption{Comparison with representative emotional support, multi-party dialogue, and mental-health support benchmarks.}
\label{tab:benchmark_comparison}
\resizebox{\linewidth}{!}{
\begin{tabular}{lcllll}
\toprule
\textbf{Benchmark} & \textbf{Year} & \textbf{Setting} & \textbf{Participants} & \textbf{Relation Signal} & \textbf{Support Signal} \\
\midrule
MELD \citep{poria2019meld}
& 2019
& Emotion recognition
& Multi-party
& Implicit speaker interaction
& None \\

MPDD \citep{chen2020mpdd}
& 2020
& Multi-party emotion dialogue
& Multi-party
& Static interpersonal relation
& None \\

ESConv \citep{liu2021towards}
& 2021
& Emotional support conversation
& One-on-one
& Not explicit
& Strategy + response \\

MPED \citep{zhu2022mped}
& 2022
& Empathetic dialogue generation
& Multi-party
& Emotion / sensibility cues
& Empathetic response \\

AugESC \citep{zheng2023augesc}
& 2023
& LLM-augmented ESC
& One-on-one
& Not explicit
& Strategy + response \\

ExTES \citep{zheng2024self}
& 2024
& LLM self-chat ESC
& One-on-one
& Not explicit
& Strategy + response \\

MESC \citep{chu2025towards}
& 2025
& Multimodal ESC
& One-on-one
& Implicit multimodal cues
& Emotion + strategy + response \\

MentalChat16K \citep{xu2025mentalchat16k}
& 2025
& Mental-health assistance
& One-on-one
& Not central
& Counseling response \\

PsyDial \citep{qiu2025psydial}
& 2025
& Long-term mental-health support
& One-on-one
& Not central
& Counseling response \\

\textbf{\toolname{} (Ours)}
& \textbf{2026}
& \textbf{Relation-aware ESC}
& \textbf{Related multi-party}
& \textbf{Directed stance + dynamic pattern}
& \textbf{Timing + target + strategy} \\

\bottomrule
\end{tabular}
}
\end{table*}

Although relation-aware emotional support has been extensively studied in psychology and psychotherapy~\citep{cox1997families,shadish2003meta,lebow2012research,joseph2025effectiveness,darwiche2026post}, which demonstrates its importance in various real-life scenarios (e.g., family~\citep{cox1997families}, couple~\citep{joseph2025effectiveness}, team~\citep{cheng2022gamification}), it remains underexplored in the AI community. Existing AI studies~\citep{gazit2025ai,wang2026simulating} have only made preliminary attempts on specific subtopics such as couple therapy, often as case studies of LLM-based relational facilitation or multi-agent therapeutic simulation. These works have not systematically examined the role of interpersonal relations in ESC, suggesting that relation-aware ESC in AI is still at an early stage. 


To address this research gap, we formulate \textbf{relation-aware ESC} as a new task that extends emotional support from individual-centered interaction to relation-centered multi-party scenarios. We instantiate this task with two representative relational scenarios, couples and families, and construct a long-duration, richly annotated benchmark named \toolname{} from real multi-party interview conversations. \toolname{} provides rich contexts in which multiple participants jointly express emotions, concerns, and interpersonal tensions. 
To contextualize \toolname{}, Table~\ref{tab:benchmark_comparison} compares it with representative benchmarks across emotional support, multi-party dialogue, and mental-health support. While prior benchmarks focus on individual support strategies~\citep{liu2021towards,zheng2023augesc,zheng2024self}, multimodal support or emotion modeling~\citep{poria2019meld,chu2025towards}, multi-party relation analysis~\citep{chen2020mpdd,zhu2022mped}, or long-term mental-health support~\citep{xu2025mentalchat16k,qiu2025psydial}, \toolname{} centers interpersonal relations in emotional support, requiring models to understand relational dynamics and make relation-sensitive support decisions.


To operationalize relation-aware ESC, we define six tasks under two dimensions: \textit{Relational Understanding} for modeling emotions, interpersonal viewpoints, and relation patterns, and \textit{Relation-Sensitive Support} for predicting intervention timing, support targets, and support strategies. By evaluating ten state-of-the-art LLMs, we find that models handle individual emotion recognition better than relation-aware reasoning. In particular, they struggle to capture relational patterns, infer interpersonal viewpoints, and make support decisions about when, whom, and how to support in multi-party scenarios.

Our main contributions are summarized as:
\begin{itemize}[leftmargin=*]
    \item We introduce relation-aware ESC, a new task that extends emotional support from individual-centered interactions to relation-centered multi-party scenarios.

    \item We construct a long-duration, richly annotated benchmark from real multi-party interview conversations in two representative relational scenarios, couples and families.

    \item We design six relation-related tasks, and benchmark state-of-the-art LLMs to reveal their limitations in modeling interpersonal relations and providing relation-sensitive support.
\end{itemize}

\section{Related Work}

\subsection{Multi-party Dialogue Generation}
Multi-party dialogue generation extends one-on-one interaction to conversations with multiple speakers, requiring models to track speaker identities, addressee relations, turn-taking, and non-linear conversational dependencies. Prior work has studied addressee and response selection~\cite{ouchi2016addressee}, neural speaker modeling~\cite{meng2018neural}, heterogeneous graph-based interaction modeling~\cite{gu2022hetermpc}, persona- and knowledge-grounded generation~\cite{ju2022personatkg}, latent addressee structures~\cite{gu2023madnet}, group-chat interaction modeling~\cite{wei2023multilight}, discourse and coherence modeling~\cite{li2024chatmdg,fan2024improving}, and LLM-based evaluation or adaptation for multi-party conversations~\cite{tan2023chatgpt,wang2024mupas}. Zhu et al.~\cite{zhu2022mped} further extend empathetic response generation to multi-party settings by modeling dynamic emotions and static speaker sensibilities. However, existing multi-party dialogue studies mainly focus on generating responses among multiple speakers, and the support scenario in Zhu et al.~\cite{zhu2022mped} is still centered on multiple responders replying to a primary help-seeker. In contrast, our task concerns multiple support recipients who are related to each other, such as parent-child or romantic partners, requiring the system to reason about their emotional needs, relational roles, and interactional tensions.

\subsection{Relation-aware Emotion Modeling}
Relation-aware emotion modeling studies how emotions in conversation are shaped by dialogue context, speaker identities, and inter-speaker dependencies, rather than by isolated utterances alone. Early datasets such as EmotionLines and MELD enable emotion analysis in multi-party conversations~\cite{chen2018emotionlines,poria2019meld}, while MPDD further incorporates interpersonal relationship annotations for studying how relations affect emotional expressions~\cite{chen2020mpdd}. Prior models track speaker-specific emotional states with recurrent architectures~\cite{majumder2019dialoguernn}, capture utterance-level dependencies with graph neural networks~\cite{ghosal2019dialoguegcn}, and model speaker and temporal relations with relation-aware graph attention~\cite{ishiwatari2020rgat}. Later work adapts pre-trained language models to multi-party emotion recognition~\cite{shen2021dialogxl}, represents conversational information flow with directed acyclic graphs~\cite{shen2021dagerc}, and incorporates external commonsense or cognitive reasoning for emotion understanding~\cite{zhong2019ket,ghosal2020cosmic,hu2021dialoguecrn}. Recent studies further explore emotion-cause reasoning and multimodal relational dependencies in conversation~\cite{kumar2023emotionflip,nguyen2024multidag}. However, these studies mainly focus on recognizing or tracking emotions. In contrast, our task requires transforming relation-aware emotional understanding into supportive responses for multiple related support recipients, where the system must balance different emotional needs, relational roles, and interactional tensions.

\section{Relation-aware Emotional Support}

\subsection{Problem Formulation}

We use a lightweight formulation to clarify the main elements evaluated in \toolname{} and how they are used to test LLMs. Consider a multi-party conversation involving a group of interrelated individuals $\mathcal{V}=\{1,\dots,n\}$. At each interaction segment $t$, the model observes a conversation context $c^t$, which includes the dialogue history and available multimodal evidence.

Each participant $i \in \mathcal{V}$ has an individual state $s_i^t$ at segment $t$, capturing their internal emotion and emotional intensity. We denote the collection of individual states as
\begin{equation}
\mathbf{s}^t = \{s_i^t\}_{i \in \mathcal{V}}.
\end{equation}

Beyond individual states, relation-aware ESC requires modeling interpersonal and group-level relational dynamics. We denote the directed interpersonal state from participant $i$ to participant $j$ as $g_{ij}^t$, and the collection of directed interpersonal states as
\begin{equation}
\mathbf{G}^t = \{g_{ij}^t\}_{i,j \in \mathcal{V}, i \neq j}.
\end{equation}
Here, $g_{ij}^t$ may include attitudes, viewpoints, alignment, or tension from $i$ toward $j$. We further denote the group-level relation pattern at segment $t$ as $\rho^t$, which summarizes the current interaction pattern among participants, such as escalation, withdrawal, repair, or alignment.

A relation-aware supporter must make support decisions based on these individual and relational states. We denote the intervention decision as
\begin{equation}
z^t \in \{0,1\},
\end{equation}
where $z^t=1$ indicates that an intervention is needed. When an intervention is made, the supporter selects a support target
\begin{equation}
\mathcal{A}^t \subseteq \mathcal{V},
\end{equation}
which may correspond to an individual, a pair, a subgroup, or the whole group, and then chooses a support strategy $a^t$.

Under this formulation, \toolname{} evaluates whether LLMs can infer individual states $(\mathbf{s}^t)$, model directed and group-level relational dynamics $(\mathbf{G}^t,\rho^t)$, and make relation-sensitive support decisions $(z^t,\mathcal{A}^t,a^t)$ from multi-party conversation contexts. These elements naturally correspond to the six benchmark tasks introduced below.

\subsection{Task Definition}

The formulation above characterizes relation-aware emotional support as a sequential decision process: the supporter first estimates evolving individual and group states, and then decides when to intervene, whom to support, and how to support them. Accordingly, we organize relation-aware ESC into two groups of observable subtasks. \textit{Relational Understanding} includes Emotion Recognition (ER), Viewpoint Prediction (VP), and Relation Pattern Prediction (RPP), which assess individual and group-state estimation. \textit{Relation-Sensitive Support} includes Intervention Time Prediction (ITP), Support Target Prediction (STP), and Support Strategy Prediction (SSP), which assess support timing, target selection, and strategy selection. Compared with traditional one-on-one ESC, which mainly involves ER and SSP~\citep{liu2021towards, zheng2023augesc, zheng2024self}, relation-aware ESC additionally requires VP, RPP, ITP, and STP to model relational dynamics and make relation-aware support decisions, as summarized in Table~\ref{tab:task_comparison}. We describe each task in detail as follows:

\providecommand{\cmark}{\ensuremath{\checkmark}}
\providecommand{\xmark}{\ensuremath{\times}}

\begin{table}[t]
\centering
\small
\resizebox{1.0\linewidth}{!}{\begin{tabular}{lcccccc}
\toprule
\textbf{Setting} 
& \textbf{ER} 
& \textbf{VP} 
& \textbf{RPP} 
& \textbf{ITP} 
& \textbf{STP} 
& \textbf{SSP} \\
\midrule
Traditional ESC & \cmark & \xmark & \xmark & \xmark & \xmark & \cmark \\
Relation-aware ESC & \cmark & \cmark & \cmark & \cmark & \cmark & \cmark \\
\bottomrule
\end{tabular}}
\caption{
Task comparison between traditional ESC and relation-aware ESC.
}
\label{tab:task_comparison}
\end{table}

\paragraph{Emotion Recognition.} We follow prior work~\citep{poria2019meld, chen2020mpdd,
ishiwatari2020rgat} to define the ER task.
Given the dialogue history and multimodal evidence of an interaction
segment, the model predicts the internal emotion and intensity of a specified
participant.

\paragraph{Viewpoint Prediction.}
The VP task evaluates whether the model can infer directed interpersonal
stance~\citep{chen2020mpdd, ishiwatari2020rgat}. Given the interaction context and the source participant, the model predicts the target participants and corresponding viewpoint descriptions.

\paragraph{Relation Pattern Prediction.}
RPP requires the model to identify the current relation pattern among
participants, such as who is dominant or vulnerable, who is aligned or
opposed, and whether the interaction is escalating, distancing, or
repairing. This task follows prior research,
which views emotional distress as shaped by recurring interpersonal
dynamics \citep{johnson2012practice, minuchin2018families}. Given the
dialogue context, the model predicts a relation-pattern label with a
brief evidence-based rationale.

\paragraph{Intervention Time Prediction.}
ITP asks whether the therapist should intervene at a candidate segment.
While one-on-one ESC typically assumes that the supporter responds after each
seeker turn \citep{liu2021towards, zheng2023augesc}, relation-aware ESC
requires timing decisions based on unfolding relational dynamics. The
model therefore predicts intervention timing by considering signals such
as escalation, withdrawal, repair attempts, or alliance rupture, which
are emphasized in therapy process and alliance research
\citep{horvath2011alliance}.

\paragraph{Support Target Prediction.}
STP predicts whom the therapist should support once an intervention is
needed. Rather than assuming a single help-seeker, the model selects the
person or relational unit that most needs support, such as one
participant, two participants in conflict, a subgroup, or the whole
group. This reflects systemic views of therapy, where distress is often
understood through relationships rather than isolated individuals
\citep{minuchin2018families, bowen1993family}.

\paragraph{Support Strategy Prediction.}
SSP predicts how the therapist should support the selected target. The strategy label captures interventions such as tracking, reframing and evoking. These strategies draw on therapy research on emotional
de-escalation, relational repair, and systemic intervention
\citep{johnson2012practice, gottman1992marital, minuchin2018families}.

Together, these tasks evaluate whether LLMs can move beyond individual
emotional support and perform relation-aware reasoning. The understanding
tasks assess participants' internal states, directed attitudes, and
relation patterns, while the support tasks assess temporally
appropriate, target-aware, and relation-sensitive intervention decisions.

\subsection{Multi-Layer Modeling}
\label{sec:multi_layer_modeling}

\begin{figure*}[htbp]
  \centering
  \includegraphics[width=1.0\linewidth]{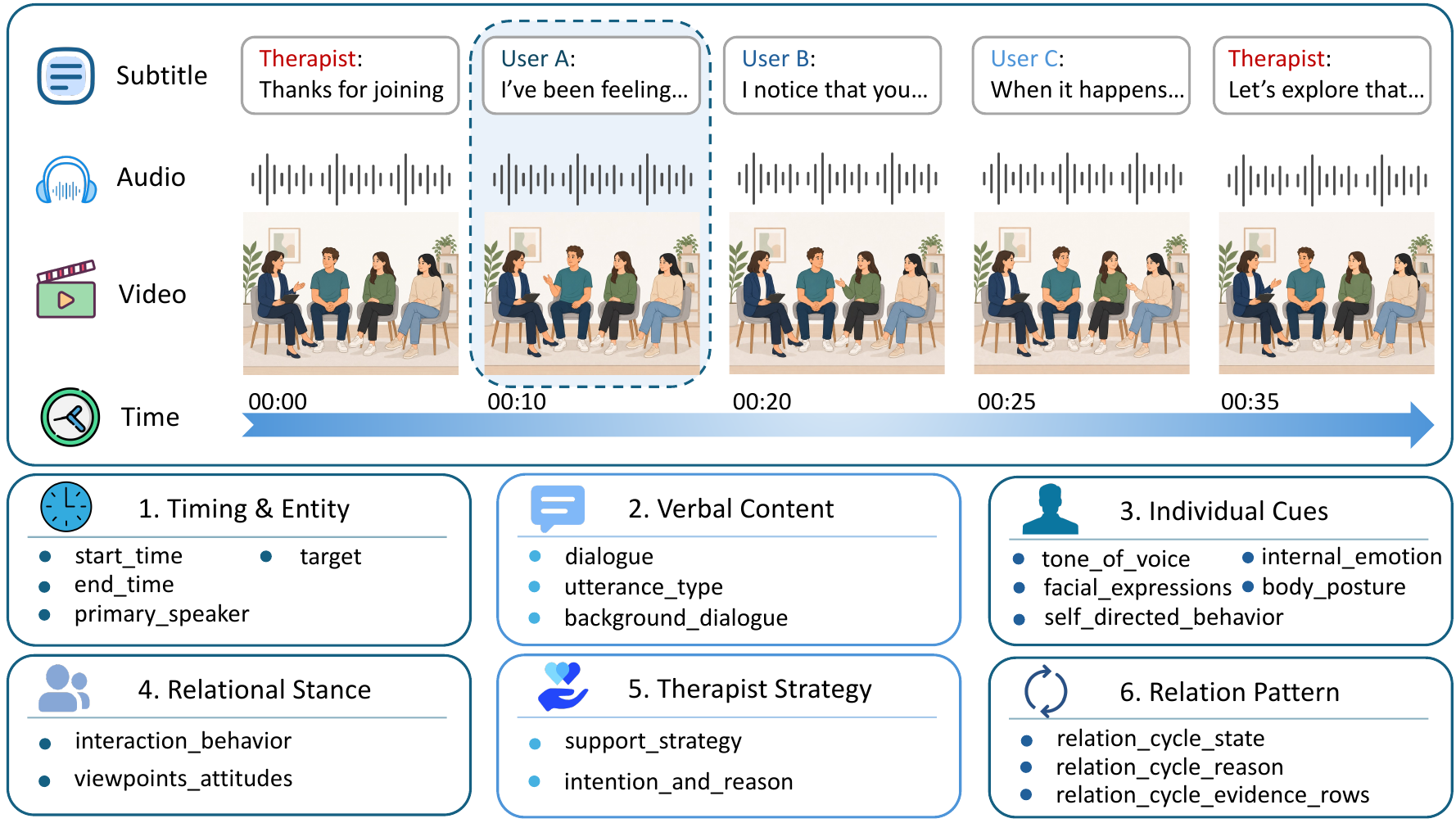}
  \caption{Overview of relational interaction dynamics modeling. We temporally align
subtitle, audio, and video streams into multimodal interaction segments.
Each segment is represented through six dimensions: timing and
entity information, verbal content, individual cues, relational stance,
therapist strategy, and relation pattern.}
  \label{fig:modeling}
\end{figure*}

To support the six benchmark tasks, we model each multi-party
conversation as a sequence of temporally grounded multimodal interaction
segments. As shown in Figure~\ref{fig:modeling}, subtitle,
audio, and video streams are aligned along a shared timeline, and each
segment is represented with six structured dimensions.

First, timing and entity information specifies the start time,
end time, primary speaker, and target, providing temporal and directed
participant grounding. Second, verbal content captures the
dialogue, utterance type, and relevant background dialogue, which provide
the semantic and conversational context of each segment. Third,
individual cues describe tone of voice, body posture, facial
expressions, self-directed behavior, and inferred internal emotion,
serving as multimodal evidence for estimating individual emotional states
$\mathit{s}^t$.

Beyond individual-level modeling, we further annotate relational and
support-related information. Relational stance captures
interaction behavior and viewpoints or attitudes toward others, providing
evidence for estimating the group states $\mathit{G}^t$. For
therapist turns, therapist strategy records the support strategy
and its intention, corresponding to the support action $a^t$ in our
formulation. Finally, relation pattern summarizes higher-level
relation-cycle states, their reasons, and supporting evidence across
segments, enabling the model to track how interpersonal dynamics evolve
over time.

Together, these dimensions bridge low-level multimodal signals and
high-level relational reasoning, supporting unified modeling of
individual emotions, interpersonal relations, therapist interventions,
and relation-cycle transitions.

\section{Dataset}
\subsection{Dataset Construction}

To facilitate the research of relation-aware ESC, we construct a benchmark from real-world multi-party interview videos. We focus on two representative relational scenarios, couples and families, and collect data from two documentary-style interview sources, \textbf{Couple Therapy} and \textbf{Family Therapy}. We manually identify independent interview segments from the videos, split them into self-contained conversation clips, and filter out clips shorter than two minutes, which usually lack sufficient relational context.

As high-quality relational annotations are essential for this complex task, we take several steps to ensure data quality. First, we use Gemini-3.1-Pro to pre-annotate each video segment according to our theoretical framework as shown in Section~\ref{sec:multi_layer_modeling}, with reference to both the video content and the aligned subtitles. Second, we build an online verification system and invite three PhD-level annotators to check the faithfulness and consistency of the annotations against the original videos. The annotators revise incorrect annotations and discard segments with severe recognition errors, speaker mismatches, or substantial inconsistency with the video evidence. Annotation details are presented in Appendix~\ref{app:annotation_procedure}. Through this process, we obtain a high-quality dataset for studying relation-aware emotional support in multi-party conversations. Per task instance construction is further detailed in Appendix~\ref{app:task_instance_construction}.

\subsection{Dataset Characteristics}
Table~\ref{tab:dataset_statistics} summarizes the dataset statistics.
The dataset contains 191 samples from two scenarios, including 174 couple
clips and 17 family clips, with 7,079 annotated turns and 1,064.8 minutes
of video in total. Family samples are longer and involve more speakers on
average, while therapist participation is also higher in family sessions
than in couple sessions.

\begin{table}[t]
\centering
\small
\setlength{\tabcolsep}{4pt}
\renewcommand{\arraystretch}{1.12}
\begin{tabular}{lccc}
\toprule
\textbf{Statistic} & \textbf{Couple} & \textbf{Family} & \textbf{All} \\
\midrule
Sample & 174 & 17 & 191 \\
Total duration (min) & 824.9 & 239.9 & 1,064.8 \\
Total turns & 5,875 & 1,204 & 7,079 \\
Avg. duration / sample & 4.74 & 14.11 & 5.57 \\
Avg. turns / sample & 33.76 & 70.82 & 37.06 \\
Avg. speakers / sample & 3.00 & 4.06 & 3.09 \\
Therapist turn share & 35.4\% & 47.1\% & 37.4\% \\
\bottomrule
\end{tabular}
\caption{Dataset statistics across two scenarios.}
\label{tab:dataset_statistics}
\end{table}





\subsection{Relational Dynamics Analysis}
\begin{figure}[t]
  \centering
  \includegraphics[width=\linewidth]{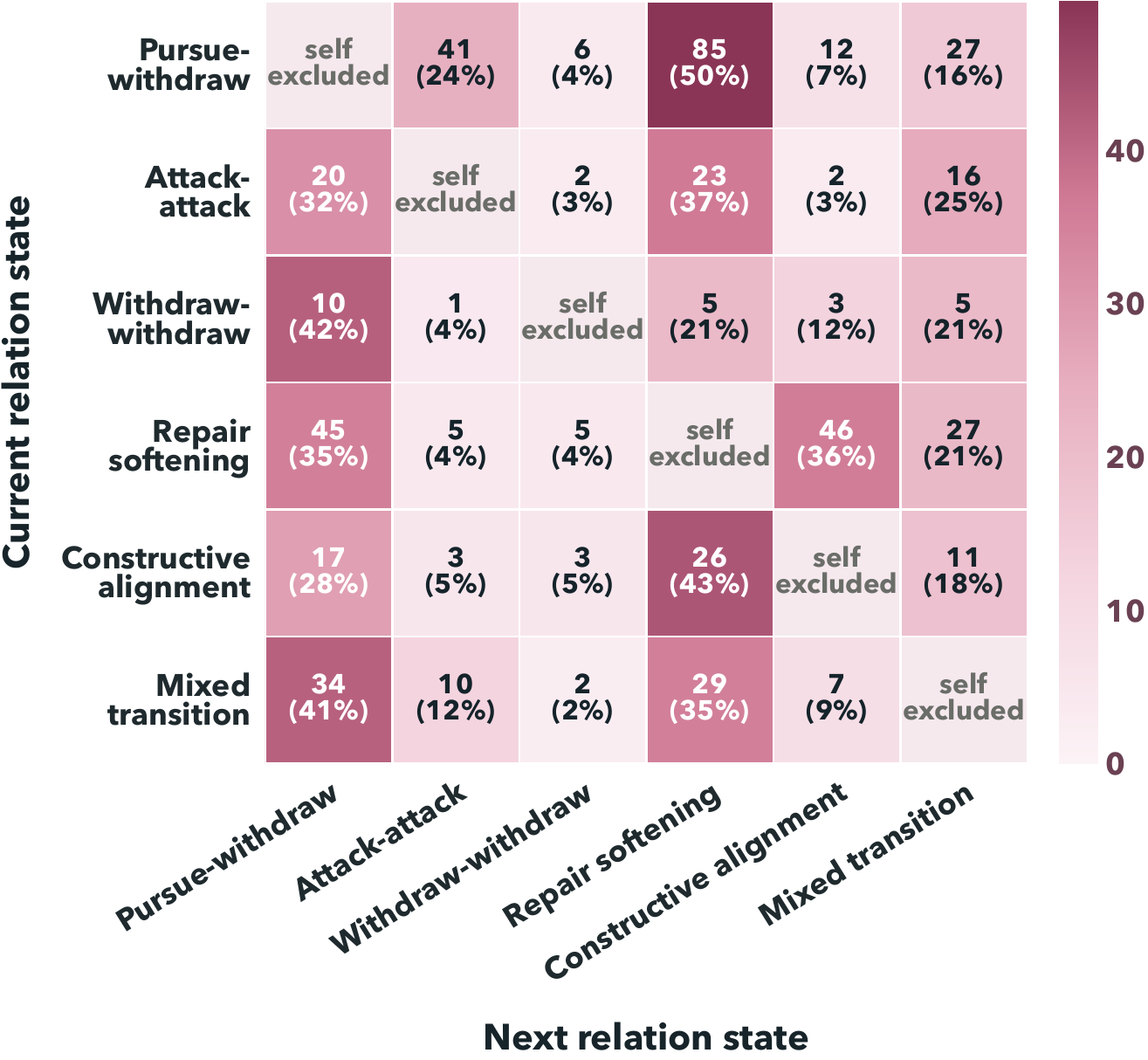}
  \caption{
  Relation-pattern transition matrix over consecutive interaction segments.
Rows denote current relation states and columns denote next relation
states. Self-transitions are excluded, and percentages are row-normalized.
  }
  \label{fig:relation_transition}
\end{figure}

To examine whether the annotated relation patterns capture meaningful
temporal dynamics, we compute a row-normalized transition matrix over
consecutive relation-pattern labels. As shown in
Figure~\ref{fig:relation_transition}, relation change is highly
nonlinear. Negative cycles such as \textit{pursue-withdraw} and
\textit{attack-attack} do not usually move directly into stable
coordination; instead, \textit{repair softening} often serves as an
intermediate state before \textit{constructive alignment}. Meanwhile,
\textit{pursue-withdraw} frequently reappears after states such as
\textit{withdraw-withdraw}, \textit{repair softening}, and
\textit{mixed transition}, suggesting that it functions as a recurring
attractor in relational interaction.

These transition patterns show that relation-aware ESC requires models
to track evolving interpersonal states rather than only recognize static
relation labels. This further motivates our relation-pattern prediction
task and provides an empirical basis for evaluating whether LLMs can
model dynamic relational change.

\section{Experiments}
Our experiments focus on two key research questions:
(1) How do existing LLMs perform on relation-aware ESC tasks?
(2) What are the key factors that lead to their success or failure?

\subsection{Model Selection}

We evaluate 10 representative LLMs: Qwen-Plus~\citep{bai2023qwen},
Qwen3-Max~\citep{yang2025qwen3}, Qwen3.5-Plus~\citep{qwen3.5}, DeepSeek-R1~\citep{guo2025deepseek}, DeepSeek-v3.2~\citep{liu2024deepseek},
DeepSeek-V4-Flash~\citep{deepseekai2026deepseekv4}, DeepSeek-V4-Pro~\citep{deepseekai2026deepseekv4}, GPT-4o~\citep{hurst2024gpt}, MiniMax M2.5, and Kimi K2.5~\citep{kimiteam2026kimik25visualagentic}.  All models are evaluated in a zero-shot setting
with the same task definitions and prompt formats (Appendix~\ref{app:task_prompt}).

\subsection{Evaluation Metrics}

We evaluate different tasks using the metrics shown in
Table~\ref{tab:evaluation_metrics}. Overall, we combine traditional
automatic metrics with LLM-based evaluation. Classification and ranking
tasks are evaluated with standard label-based metrics, while generative
understanding tasks are evaluated using GPT-5.4 as an LLM-as-judge with a
5-point Likert scale, together with BERTScore for semantic similarity. 
We further validate the correlation between human and LLMs through human evaluation (Appendix~\ref{app:human_eval}).

\begin{table}[t]
\centering
\small
\setlength{\tabcolsep}{4pt}
\renewcommand{\arraystretch}{1.15}
\begin{tabular}{lll}
\toprule
\textbf{Task} & \textbf{Type} & \textbf{Metrics} \\
\midrule
ITP 
& Binary classification 
& Prec., Rec., F1 \\

RPP
& Multiclass classification 
& Acc. \\

STP 
& Ranking 
& Rec., MRR \\

SSP 
& Ranking 
& Rec., MRR \\

ER 
& Generation 
& LLM-as-judge, BERTScore \\

VP 
& Generation 
& LLM-as-judge, BERTScore \\
\bottomrule
\end{tabular}

\caption{Evaluation metrics for different relation-aware ESC tasks.}
\label{tab:evaluation_metrics}
\end{table}

\subsection{Results and Analysis}

\paragraph{Main Results.}
Table~\ref{tab:model-performance} reports the performance of 10 representative LLMs on the proposed relation-aware ESC benchmark. Overall, current LLMs perform reasonably well on tasks that rely more on local intervention or affective cues, such as ITP and ER, where the average ITP F1 reaches 82.62\% and the average ER LLM-as-judge score reaches 4.05/5. This suggests that existing models can often identify emotionally salient moments and infer individual affective states from dialogue context.

\begin{table*}[t]
\centering
\small
\setlength{\tabcolsep}{3.2pt}
\renewcommand{\arraystretch}{0.92}
\resizebox{\linewidth}{!}{\begin{tabular}{lccc c cc cc cc cc}
\toprule
\multirow{2}{*}{Model}
& \multicolumn{3}{c}{ITP}
& \multicolumn{1}{c}{RPP}
& \multicolumn{2}{c}{STP}
& \multicolumn{2}{c}{SSP}
& \multicolumn{2}{c}{ER}
& \multicolumn{2}{c}{VP} \\
\cmidrule(lr){2-4}
\cmidrule(lr){5-5}
\cmidrule(lr){6-7}
\cmidrule(lr){8-9}
\cmidrule(lr){10-11}
\cmidrule(lr){12-13}
& Prec. & Rec. & F1
& Acc.
& Rec. & MRR
& Rec. & MRR
& LLM & BERT
& LLM & BERT \\
\midrule
\texttt{Qwen-Plus} & 55.78 & 89.99 & 68.87 & 38.27 & 65.61 & 79.21 & 26.21 & 36.95 & 3.58 & 0.8167 & 3.20 & 0.8587 \\
\texttt{Qwen3-Max} & 79.81 & 70.99 & 75.14 & 43.22 & 61.09 & 76.28 & 31.89 & 45.76 & 4.04 & 0.8224 & 3.35 & 0.8647 \\
\texttt{Qwen3.5-Plus} & 98.00 & 91.42 & 94.60 & 40.27 & 69.40 & 81.57 & 31.70 & 44.98 & 4.22 & 0.8321 & 3.94 & 0.8642 \\
\texttt{DeepSeek-R1} & 90.71 & 72.12 & 80.36 & 37.89 & 61.88 & 76.88 & 29.38 & 39.71 & 4.08 & 0.8227 & 3.35 & 0.8635 \\
\texttt{DeepSeek-V3.2} & 93.42 & 81.17 & 86.86 & 36.08 & 62.68 & 77.14 & 33.05 & 46.17 & 4.13 & 0.8259 & 3.86 & 0.8588 \\
\texttt{DeepSeek-V4-Flash} & 91.10 & 78.31 & 84.22 & 39.39 & 62.61 & 77.61 & 30.12 & 44.13 & 4.04 & 0.8153 & 4.08 & 0.8577 \\
\texttt{DeepSeek-V4-Pro}   & 92.24 & 75.91 & 83.28 & 45.60 & 71.04 & 82.23 & 37.26 & 52.00 & 4.17 & 0.8246 & 3.90 & 0.8574 \\
\texttt{Kimi K2.5} & 88.01 & 84.03 & 85.98 & 45.26 & 59.13 & 75.60 & 34.15 & 47.66 & 4.18 & 0.8225 & 3.64 & 0.8620 \\
\texttt{MiniMax M2.5} & 89.64 & 75.98 & 82.25 & 39.85 & 62.55 & 77.01 & 29.08 & 42.89 & 4.01 & 0.8201 & 3.25 & 0.8561 \\
\texttt{GPT-4o} & 86.67 & 82.63 & 84.60 & 38.69 & 67.38 & 80.00 & 35.92 & 49.36 & 4.01 & 0.8206 & 3.24 & 0.8674 \\
\bottomrule
\end{tabular}}
\caption{Model performance across relation-aware ESC tasks.}
\label{tab:model-performance}
\end{table*}

Performance drops substantially when the task requires explicit relational reasoning. RPP is particularly challenging, with the best accuracy only reaching 45.60\% and the model average being 40.45\%. A similar gap appears in VP: although its average BERT score is relatively high (0.8611), the average LLM-as-judge score is notably lower than ER (3.58/5 vs.\ 4.05/5), suggesting that inferring viewpoints toward another person is harder than recognizing one's own emotion. The support-side tasks further reflect this limitation. Performance is moderate on STP (64.34\% recall, 78.35\% MRR), but drops further on SSP (31.88\% recall, 44.96\% MRR), indicating that determining an appropriate relation-sensitive support strategy is particularly difficult.

Among all models, \texttt{Qwen3.5-Plus} achieves the strongest performance on ITP and ER, with 94.60\% ITP F1 and a 4.22 ER score, but its advantage is inconsistent on relation-intensive tasks. Its RPP accuracy is only 40.27\%, below \texttt{DeepSeek-V4-Pro}, \texttt{Kimi K2.5}, and \texttt{Qwen3-Max}, and it does not achieve the best results on STP, SSP, or VP. Instead, \texttt{DeepSeek-V4-Pro} performs best on RPP, STP, and SSP, while \texttt{DeepSeek-V4-Flash} obtains the highest VP LLM-as-judge score. These results suggest that even strong general-purpose LLMs still struggle to move from individual-level understanding to robust relation-level reasoning and support planning. Results under the fine-grained scenario categories are provided in Appendix~\ref{app:scenario_based_results}.

\paragraph{Failure Analysis.}
Since models perform worse on relation-related tasks, we select three representative tasks, RPP, VP, and SSP, for failure analysis.

\begin{figure}[t]
  \centering
  \includegraphics[width=\linewidth]{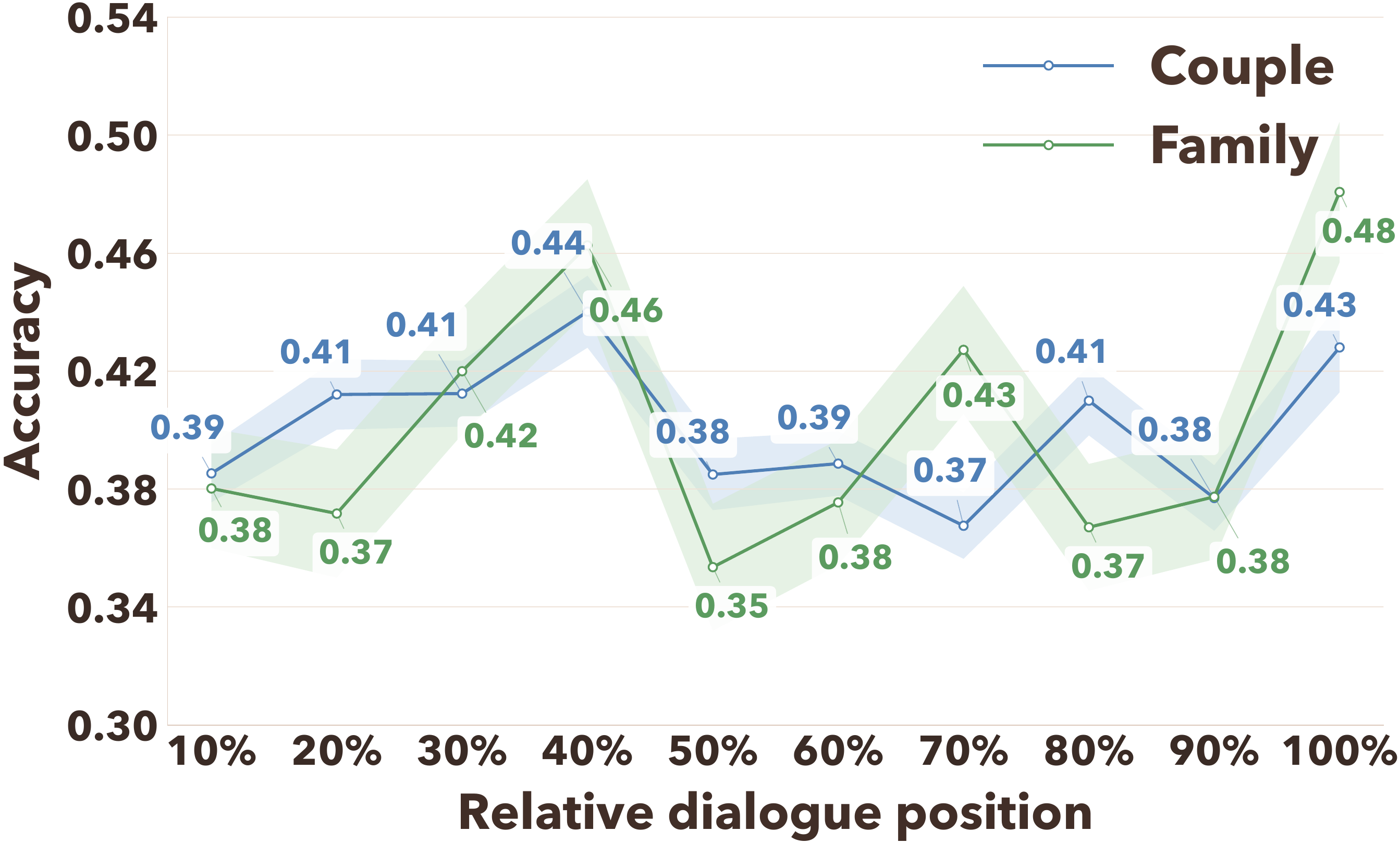}
  \caption{Accuracy of the RPP task across relative dialogue position. Each dialogue is normalized into ten position bins (0-10\% to 90-100\%), and results are pooled across all evaluated models. Shaded regions denote standard error.}
  \label{fig:rpp_distribution}
\end{figure}
\begin{figure}[t]
  \centering
  \includegraphics[width=\linewidth]{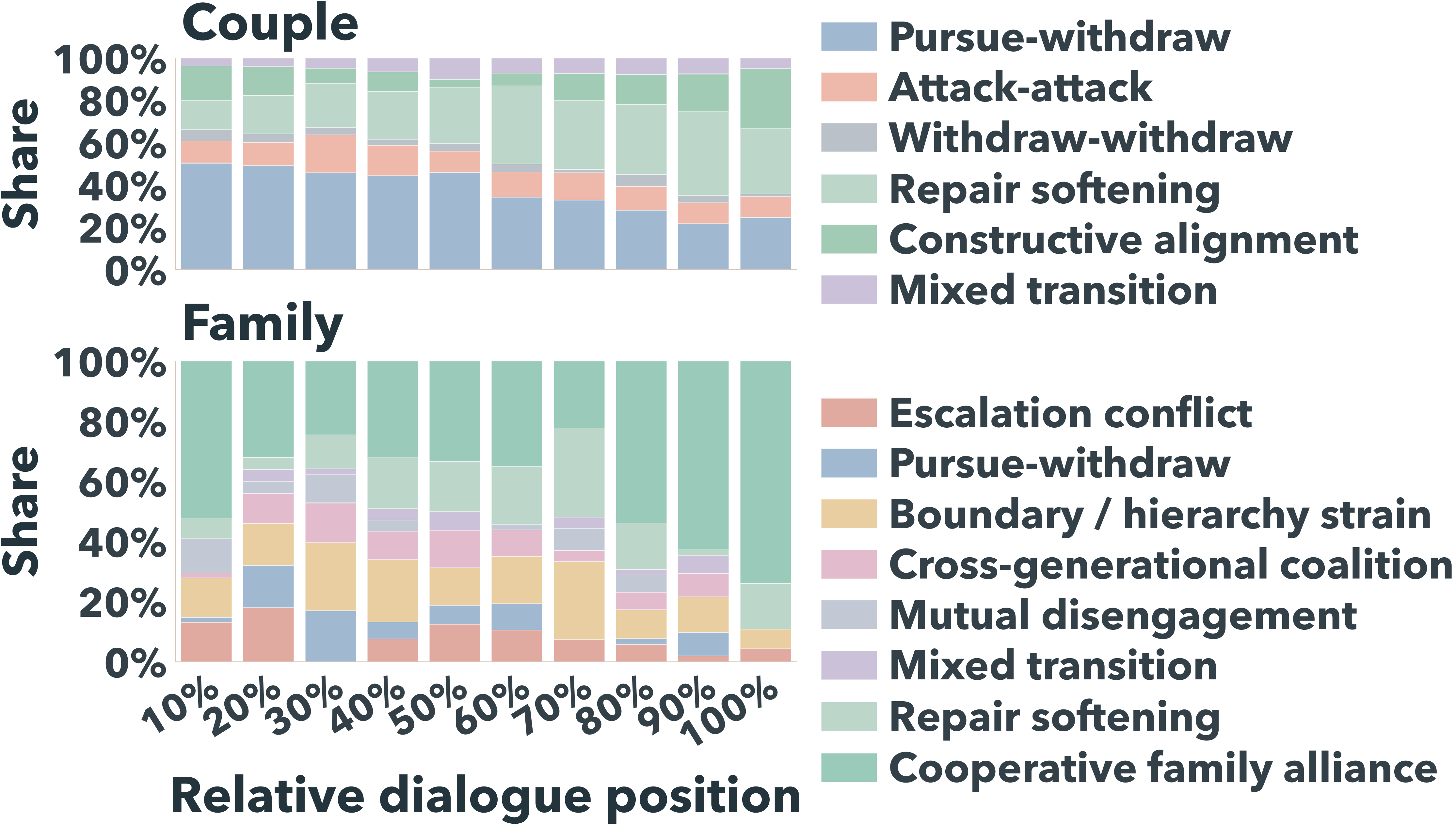}
  \caption{Gold relation-pattern distribution across relative dialogue position. Bars indicate the proportion of annotated relation patterns within each position bin.}
  \label{fig:relation_pattern_distribution}
\end{figure}

We first analyze RPP performance across dialogue stages. As shown in Figure~\ref{fig:rpp_distribution}, accuracy fluctuates rather than declining monotonically as the dialogue progresses. This suggests that the main challenge of RPP lies not simply in handling longer context, but in tracking the dynamic transition of relation patterns over time.
Figure~\ref{fig:relation_pattern_distribution} provides further evidence: relation-pattern distributions shift across dialogue stages, leading to substantial variation in prediction accuracy. This requires models to continuously update their understanding of evolving interpersonal structure as relation-pattern distributions shift over time.

For VP, the main difficulty comes from the indirect expression of directed viewpoints. Participants do not always address each other directly; instead, they convey their attitudes through the therapist, and such expressions are often implicit. For example, vulnerable feelings such as hurt or grievance may be disguised as anger or dominance. Moreover, models tend to capture only coarse-grained viewpoints, but fail to perform fine-grained reasoning based on the relational context. As a result, they often misinterpret the target-specific viewpoints behind a participant's utterance.

For SSP, the low recall suggests that models often fail to precisely identify the optimal support strategy. However, the relatively higher MRR indicates that they can still rank plausible strategies near the top. In other words, models can roughly narrow down the candidate strategy set, but struggle with fine-grained strategy selection.

\section{Conclusion}
In this paper, we introduced relation-aware emotional support conversation, a new task that extends traditional ESC to relation-centered multi-party scenarios. We constructed \toolname{} from real couple and family interview conversations and defined six tasks covering both relational understanding and support decision-making.
Experiments with ten LLMs show that current models still struggle with relation-intensive tasks, including relation pattern prediction, viewpoint prediction, and support strategy selection. These findings reveal the limitations of current LLMs in modeling interpersonal relations and making support decisions grounded in relational dynamics.
Moreover, we hope \toolname{} will encourage future research toward more relation-sensitive and context-grounded emotional support systems.

\section{Acknowledgment}
We thank the anonymous area chair and anonymous
reviewers for their insightful comments and valuable feedback during the review process. This study
is funded by the Research Grants Council (project
code: T43-518/24-N and PolyU/15213323) under
the University Grants Committee, Hong Kong Special Administrative Region Government.

\section*{Limitations}

Despite these contributions, we identify four main limitations of this work. 

First, our benchmark is constructed from publicly available documentary-style couple and family interview videos. Although such data may reflect selection biases introduced by media production, editing, participant demographics, and cultural context, they also provide rich, ecologically meaningful relational interactions that are difficult to capture in controlled laboratory settings. Future work could further improve demographic and cultural diversity by incorporating broader sources of naturally occurring relational interactions. 

Second, relation-aware emotional support involves inherently subjective judgments, especially for high-level labels such as relation patterns, directed viewpoints, support targets, and support strategies. While such subjectivity cannot be fully eliminated, our LLM-assisted pre-annotation followed by expert human verification provides a practical and scalable way to improve annotation consistency. Future work could further strengthen reliability through larger annotator pools and more fine-grained annotation guidelines. 

Third, several tasks exhibit long-tailed label distributions, which may affect both model training and evaluation, especially for rare relation patterns or support strategies. 
Nevertheless, these long-tailed distributions also reflect the natural imbalance of real-world relational and supportive behaviors, making the benchmark more realistic. Future studies could explore data augmentation, rebalancing strategies, or rare-label evaluation protocols to better address this issue. 

Fourth, due to copyright and privacy considerations, we do not redistribute raw videos, audio, or visual content, which may limit full multimodal reproducibility for researchers without access to the original sources. However, this decision helps ensure ethical data use and protects the rights and privacy of individuals appearing in the source materials. Future work could investigate privacy-preserving data-sharing mechanisms or controlled-access protocols to improve reproducibility while maintaining ethical safeguards. 

Despite these limitations, we believe this benchmark provides a valuable foundation for future research on relation-aware emotional support in realistic couple and family interactions.

\section*{Ethics Statement}

\paragraph{Source material and copyright.}
Our benchmark is constructed from publicly available documentary-style couple\footnote{\url{https://www.youtube.com/@couplestherapySHO}} and family\footnote{\url{https://www.youtube.com/playlist?list=PLedeM8OP4sIlAjz_wLzlvYaNZ3XqYlbrm}} interview videos. We use these videos only as source material for academic research on relation-aware emotional support conversations. The original videos, audio streams, subtitles, screenshots, and other copyrighted media assets are not redistributed as part of our benchmark. Instead, our released resources focus on derived annotations and task instances. When necessary, we provide source identifiers and temporal metadata that allow researchers to locate the corresponding public materials, subject to the availability and licensing terms of the original sources.

\paragraph{Data release.}
We plan to release RESCUE-BENCH under a research-only license. The public release will include derived task data, speaker identifiers, label taxonomies, task definitions, evaluation scripts, prompts, and aggregate statistics. To support reproducibility while respecting copyright and participant privacy, we will not publicly redistribute copyrighted raw videos, raw audio, screenshots, or unrestricted full transcripts when such redistribution is not permitted by the original source licenses. Instead, where legally and ethically permissible, we will provide limited controlled access to the necessary raw audio/video materials or minimally sufficient source snippets on a case-by-case basis. Access will be granted only to qualified researchers who provide a clear research purpose, institutional affiliation, identity credentials, and, where applicable, IRB approval or ethics exemption. Approved users will be required to sign a data use agreement that prohibits re-identification, redistribution of source media, participant profiling, attempts to recover private identities, commercial use, and clinical deployment. Sensitive or potentially identifying derived content will also be restricted to research use under the same terms. This controlled-access protocol is intended to balance reproducibility with the legal and ethical constraints of using real-world therapeutic and interview-style recordings.

\paragraph{Annotator qualifications and compensation.}
All annotations are conducted by trained annotators with relevant backgrounds in psychology. Before annotation, annotators are provided with task-specific guidelines to ensure consistent interpretation. Given the potentially sensitive nature of the materials, annotators are instructed to approach the data with care and to avoid making clinical diagnoses or judgments about the individuals represented. Annotators are compensated at a reasonable rate for their work, in accordance with the expected expertise, time commitment, and complexity of the annotation tasks.

\paragraph{Intended use.}
This benchmark is intended solely for research purposes, specifically for evaluating model capabilities in relation-aware emotional support conversation, including relational understanding and support planning. It is not designed for clinical deployment and should not be used as a clinical decision-making system, a substitute for professional therapy, or a tool for diagnosing, assessing, or evaluating real individuals. As the benchmark is derived from documentary-style public videos, researchers should use the data in a manner that respects the dignity, privacy, and contextual integrity of the individuals represented. Any model outputs or performance results obtained from this benchmark should be interpreted responsibly and should not be used to make consequential judgments about real people, their emotions, or their relationships.

\paragraph{Use of AI assistants.}
We used AI assistants such as ChatGPT for writing assistance, including language polishing and clarity improvement. All AI-generated content was carefully reviewed, verified, and revised by the authors, who take full responsibility for the final content of the paper.

\bibliography{custom}

\clearpage
\appendix
\startcontents[appendices]
\printcontents[appendices]{}{1}{\section*{Appendix Contents}}

\section{Detailed Task Definitions}
\label{app:task_definitions}

This section provides detailed definitions of the six benchmark tasks in relation-aware emotional support conversation.
All tasks are built on temporally ordered multi-party therapy conversations.
Given the dialogue context up to a certain point, the model is required to either understand the current client-side state or predict the therapist's next support decision.
Future turns are not visible to the model during prediction.

\subsection{Intervention Timing Prediction}
\label{app:itp}

\paragraph{Task.}
Intervention Timing Prediction (ITP) evaluates whether the therapist should intervene at a candidate point in the conversation.

\paragraph{Input.}
The input is the conversation history up to a candidate boundary, together with the current client-side turn after which an intervention decision is considered.

\paragraph{Output.}
The output is a binary intervention decision, \texttt{should\_speak} $\in \{\texttt{yes}, \texttt{no}\}$.
A prediction of \texttt{yes} means that the therapist should speak immediately after the current boundary, while \texttt{no} means that the therapist should not intervene at this point.

\paragraph{Evaluation.}
ITP is evaluated as a binary classification task.
We treat \texttt{yes} as the positive class and report Precision, Recall, and F1:
$P=\frac{TP}{TP+FP}$, $R=\frac{TP}{TP+FN}$, and $F1=\frac{2PR}{P+R}$.

\subsection{Support Target Prediction}
\label{app:stp}

\paragraph{Task.}
Support Target Prediction (STP) evaluates whom the therapist should primarily support in the next therapist intervention.

\paragraph{Input.}
The input is the conversation history before a therapist intervention, together with the set of possible support targets.
The target may be an individual participant or a relational unit, such as the couple or the family as a whole.

\paragraph{Output.}
The output is a ranked list of three candidate support targets, denoted as \texttt{target\_top3}.
Each item is selected from the candidate target set.
The first-ranked target is the model's primary prediction.

\paragraph{Evaluation.}
STP is evaluated as a top-3 ranking task.
In the main results, \textbf{Rec.} denotes Recall@1, i.e., whether the gold support target is ranked first.
MRR is computed based on the gold target's rank within the top-3 list: if the gold target appears at rank $r \in \{1,2,3\}$, its reciprocal rank is $1/r$; otherwise it is 0.
We report the average Recall@1 and MRR over all instances.

\subsection{Support Strategy Prediction}
\label{app:ssp}

\paragraph{Task.}
Support Strategy Prediction (SSP) evaluates how the therapist should support a given target in the next intervention.
In contrast to STP, the support target is provided as part of the input, so the model focuses on selecting an appropriate relation-sensitive support strategy.

\paragraph{Input.}
The input is the conversation history before a therapist intervention, together with the known support target.
In some instances, the recent internal emotional state of the support target is also provided as additional context.

\paragraph{Output.}
The output is a ranked list of three support strategy labels, denoted as \texttt{strategy\_top3}.
Each item is selected from the predefined strategy label set, such as validation, reframing, de-escalation, perspective-taking, boundary clarification, or repair guidance.
The first-ranked strategy is the model's primary prediction.

\paragraph{Evaluation.}
SSP is evaluated as a top-3 ranking task.
In the main results, \textbf{Rec.} denotes Recall@1, i.e., whether the gold support strategy is ranked first.
MRR is computed based on the gold strategy's rank within the top-3 list: if the gold strategy appears at rank $r \in \{1,2,3\}$, its reciprocal rank is $1/r$; otherwise it is 0.
We report the average Recall@1 and MRR over all therapist intervention instances.

\subsection{Relation Pattern Prediction}
\label{app:rpp}

\paragraph{Task.}
Relation Pattern Prediction (RPP) evaluates whether the model can identify the dominant relation pattern at a target moment.
In our benchmark, relation patterns are operationalized as relation-cycle states, capturing recurring interpersonal dynamics such as pursue-withdraw, attack-attack, mutual disengagement, repair softening, or constructive alignment.

\paragraph{Input.}
The input is the conversation context before the target moment, optionally including the current target turn.
The model is also given the candidate relation-pattern labels.

\paragraph{Output.}
The output is a relation-pattern label selected from the candidate label set.
When the model produces a ranked list, the top-ranked label is used as the final prediction.

\paragraph{Evaluation.}
RPP is evaluated as a multiclass classification task.
We report Accuracy, computed as the proportion of instances where the predicted relation-pattern label matches the gold label.

\subsection{Emotion Recognition}
\label{app:er}

\paragraph{Task.}
Emotion Recognition (ER) evaluates whether the model can infer the current speaker's internal emotional state.
This task focuses on the speaker's self-state rather than their attitude toward another participant.

\paragraph{Input.}
The input is the current client turn, optionally together with its preceding dialogue context and multimodal cues.

\paragraph{Output.}
The output contains a short natural-language description of the speaker's internal emotion and an intensity score.
The emotion description captures what the speaker internally feels at the current moment, while the intensity score indicates the strength of that emotion.

\paragraph{Evaluation.}
ER is evaluated as a generative understanding task.
For the emotion description, we use an LLM-as-judge score
$S_{\mathrm{emo}} \in \{1,2,3,4,5\}$ to measure semantic consistency between the predicted and gold internal emotion descriptions, where a higher score indicates better semantic alignment.
For the intensity score, we compare the predicted intensity $\hat{y}_{\mathrm{int}}$ with the gold intensity $y_{\mathrm{int}}$, both on a 0--10 scale.
The intensity matching score is computed as:
\begin{equation}
S_{\mathrm{int}}
=
\max\left(1,\; 5 - |\hat{y}_{\mathrm{int}} - y_{\mathrm{int}}|\right),
\end{equation}
where an exact match receives 5 points, each one-point difference reduces the score by 1, and differences of 4 points or more receive 1 point.
The final ER score is the average of the emotion and intensity scores:
\begin{equation}
S_{\mathrm{ER}}
=
\frac{S_{\mathrm{emo}} + S_{\mathrm{int}}}{2}.
\end{equation}
We additionally report BERTScore F1 between the predicted and gold emotion descriptions as a semantic similarity metric.

\subsection{Viewpoint Prediction}
\label{app:vp}

\paragraph{Task.}
Viewpoint Prediction (VP) evaluates whether the model can infer directed interpersonal viewpoints expressed in the current client turn.
While ER asks what the speaker feels internally, VP asks what the speaker believes, assumes, or expresses about another participant.

\paragraph{Input.}
The input is the current client turn, optionally together with its preceding dialogue context and multimodal cues.

\paragraph{Output.}
The output is a structured list of directed viewpoints.
Each viewpoint consists of a source speaker, a target participant, and a concise natural-language description of the viewpoint.
If the current turn expresses multiple distinct viewpoints toward different participants, the model may output multiple source-target-viewpoint triples.

\paragraph{Evaluation.}
VP is evaluated as a generative understanding task.
We report an LLM-as-judge score for the predicted viewpoints and BERTScore for semantic similarity between the predicted and gold viewpoint descriptions.

\section{A Sequential Decision View of Relation-aware ESC}
\label{app:sequential_view}

In the main text, we use a lightweight benchmark-oriented formulation to define the observable elements evaluated in \toolname{}. Here, we provide a broader sequential decision view of relation-aware emotional support. This view is not used as an optimization objective in this work; rather, it illustrates how future relation-aware ESC systems may explicitly model the long-term effects of support actions on both individual emotional states and interpersonal relations.

Consider a group of interrelated individuals $\mathcal{V}=\{1,\dots,n\}$. At interaction segment $t$, each participant $i \in \mathcal{V}$ has an individual state $s_i^t$, which captures their internal emotion and emotional intensity. The collection of individual states is denoted as
\begin{equation}
\mathbf{s}^t
=
\{s_i^t\}_{i\in\mathcal{V}}.
\end{equation}
In addition, the directed relational state from participant $i$ to participant $j$ is denoted as $g_{ij}^t$. Rather than assuming that this state is a scalar value, we treat $g_{ij}^t$ as a structured relational representation that may include viewpoints, attitudes, alignment, or tension from $i$ toward $j$. The collection of directed relational states is denoted as
\begin{equation}
\mathbf{G}^t
=
\{g_{ij}^t\}_{i,j\in\mathcal{V}, i\neq j}.
\end{equation}
The combined state is then
\begin{equation}
\mathbf{x}^t
=
(\mathbf{s}^t,\mathbf{G}^t).
\end{equation}

\paragraph{Target Selection and Support Strategy.}
At each step, the supporter selects a subset of recipients,
\begin{equation}
\mathcal{A}^t
\subseteq
\mathcal{V},
\end{equation}
where $\mathcal{A}^t$ may correspond to an individual, a pair, a subgroup, or the whole group. The supporter then generates a support action according to a relation-aware support strategy:
\begin{equation}
a^t
=
\sigma(
    \mathbf{x}^t,
    \mathcal{A}^t
).
\end{equation}
Here, $\sigma$ maps the current individual and relational states, together with the selected support target, to a support action. Different from relation-agnostic support, the strategy explicitly considers both individuals' internal emotional states and their interpersonal relations.

\paragraph{Joint Update of Individual and Relational States.}
A support action may influence both individual emotional states and relational states. Thus, the group under support may evolve according to
\begin{equation}
\begin{aligned}
\mathbf{s}^{t+1}
&=
f_s(
    \mathbf{x}^t,
    \mathcal{A}^t,
    a^t
), \\
\mathbf{G}^{t+1}
&=
f_G(
    \mathbf{x}^t,
    \mathcal{A}^t,
    a^t
).
\end{aligned}
\end{equation}
Equivalently,
\begin{equation}
(\mathbf{s}^{t+1},\mathbf{G}^{t+1})
=
F(
    \mathbf{x}^t,
    \mathcal{A}^t,
    a^t
).
\end{equation}

\paragraph{Conceptual Group-level Objective.}
Conceptually, an ideal relation-aware support strategy should improve group-level well-being while avoiding neglecting the most distressed participant. This can be expressed as
\begin{equation}
\begin{aligned}
\sigma^\star
=
\arg\max_{\sigma}
\;
\mathbb{E}_{\sigma}
\Bigg[
\sum_{t=0}^{T}
\gamma^t
\Bigg(
&U(\mathbf{s}^t,\mathbf{G}^t)
-
\lambda
\max_{i\in\mathcal{V}}
d(s_i^t)
\Bigg)
\Bigg].
\end{aligned}
\end{equation}
Here, $\gamma\in[0,1]$ is a discount factor, $U(\mathbf{s}^t,\mathbf{G}^t)$ denotes a conceptual group-level utility that depends on both individual emotional states and interpersonal relations, and $d(s_i^t)$ denotes the distress level of participant $i$. The regularization term penalizes leaving the most distressed participant unsupported, and $\lambda$ controls the strength of this penalty.

We emphasize that \toolname{} does not instantiate or optimize this objective. Instead, it evaluates observable components of this sequential decision process through six benchmark tasks. This sequential decision view provides a foundation for future work on optimizing relation-aware support agents, where agents may learn to improve their intervention policies by explicitly modeling the long-term effects of support actions on both individual emotional states and interpersonal relations.

\section{Task Instance Construction}
\label{app:task_instance_construction}

This section describes how we construct task instances from the row-level annotated conversations. The original data directory contains one empty manifest JSON file, which is excluded from all task construction. After removing this invalid file, the benchmark contains 174 couple samples and 17 family samples.

During construction, we filter out rows or checkpoints that do not contain the required gold annotations, whose speaker or target cannot be resolved, whose labels fall outside the corresponding task taxonomy, whose labels are empty because annotators could not identify a clear label from the available evidence, or whose dialogue content is too short or invalid for the target task. Table~\ref{tab:task_construction_stats} summarizes the final number of retained instances and covered samples for each task.

\begin{table*}[t]
\centering
\small
\begin{tabular}{@{}l>{\raggedright\arraybackslash}p{0.42\textwidth}rrrr@{}}
\toprule
Task & Construction Rule & Couple Inst. & Family Inst. & Couple Samp. & Family Samp. \\
\midrule
ITP & Balanced binary decision points after dialogue boundaries
& 2722 & 284 & 168 & 16 \\
RPP & Rows with non-empty gold relation-pattern label
& 1608 & 525 & 172 & 17 \\
STP & Therapist rows with valid support target
& 1290 & 347 & 172 & 17 \\
SSP & Therapist rows with non-empty strategy and resolvable target
& 1290 & 347 & 172 & 17 \\
ER & Client-speaking checkpoints with valid emotion annotation
& 844 & 82 & 174 & 17 \\
VP & Client-speaking checkpoints with valid viewpoint annotation
& 775 & 80 & 170 & 17 \\
\bottomrule
\end{tabular}
\caption{Task instance construction statistics. ``Inst.'' denotes the final number of task instances after filtering invalid rows or checkpoints, and ``Samp.'' denotes the number of covered samples.}
\label{tab:task_construction_stats}
\end{table*}

\paragraph{Intervention Time Prediction (ITP).}
For ITP, each instance is a candidate decision point. We traverse dialogue boundaries in each sample and construct positive and negative candidates. A positive candidate is retained when the current row is spoken by the therapist and the previous row is spoken by a non-therapist participant, corresponding to an actual therapist entry point. A negative candidate is retained when both the current row and the next row are spoken by non-therapist participants, indicating that the therapist should continue listening. We discard samples that do not contain valid positive or negative candidates, and balance positive and negative candidates within each sample. This yields 2,722 couple instances and 284 family instances, covering 168 couple samples and 16 family samples.

\paragraph{Relation Pattern Prediction (RPP).}
For RPP, each instance is a target row with a concrete gold relation-pattern label. We traverse all rows and retain rows whose relation-pattern label is non-empty and included in the task taxonomy. Rows are discarded if they do not contain a relation-pattern annotation, if their label is outside the taxonomy, or if the label is empty because annotators judged that no clear dominant relation pattern could be identified from the available evidence. Unlike support-side tasks, RPP is not restricted to therapist turns; it covers all rows with valid non-empty relation-pattern labels. This yields 1,608 couple instances and 525 family instances, covering 172 couple samples and all 17 family samples.

\paragraph{Support Target Prediction (STP).}
For STP, each instance is a therapist turn with a valid support target. We first select rows whose primary speaker is the therapist, and then extract the gold support target from the row-level annotation. Rows are discarded if the target is missing, empty, or cannot be resolved to the closed target set of the corresponding sample. Each retained row forms a ranking instance over candidate support targets. This yields 1,290 couple instances and 347 family instances, covering 172 couple samples and all 17 family samples.

\paragraph{Support Strategy Prediction (SSP).}
For SSP, each instance is also a therapist turn, but it must contain both a concrete gold support strategy and a resolvable support target. We discard rows whose strategy is missing, empty because annotators could not determine a clear strategy, outside the scenario-specific taxonomy, or not grounded to a valid target. When the annotated support target is missing, we fall back to the row-level target field; if the target still cannot be resolved, the row is removed. This yields 1,290 couple instances and 347 family instances, covering 172 couple samples and all 17 family samples.

\paragraph{Emotion Recognition (ER).}
For ER, each instance is a sampled checkpoint from a client-speaking row. We exclude rows spoken by the therapist or by group-level speakers such as \texttt{Couple} or \texttt{Family}. We retain a row only if the current speaker has a speaker-matched gold emotion label and a valid intensity value. Rows with very short or invalid dialogue content are removed. After filtering valid candidates, we discard early candidates within each sample and uniformly sample at most five checkpoints from the remaining candidates. This yields 844 couple instances and 82 family instances, covering all 174 couple samples and all 17 family samples.

\paragraph{Viewpoint Prediction (VP).}
For VP, each instance is a sampled checkpoint from a client-speaking row with a valid directed viewpoint annotation. We exclude therapist and group-level speaker rows. We retain a row only if its \texttt{viewpoints\_attitudes} field contains at least one gold viewpoint whose source matches the current speaker, and the viewpoint has both a non-empty target and a non-empty viewpoint description. Rows without a speaker-matched viewpoint, a resolvable target, or valid viewpoint content are discarded. After filtering valid candidates, we uniformly sample at most five checkpoints from each sample. This yields 775 couple instances and 80 family instances, covering 170 couple samples and all 17 family samples.

Overall, ITP, RPP, STP, and SSP are constructed by traversing eligible dialogue boundaries or annotated rows, while ER and VP are constructed by filtering valid client-speaking checkpoints and then sampling a small number of checkpoints from each sample. The final instance counts differ across tasks because each task requires different gold annotations and applies different validity constraints. For RPP and SSP, rows with empty labels are excluded from the final task instances, since empty labels indicate that annotators could not identify a clear relation pattern or support strategy from the available evidence.
\section{Label Taxonomy}
\label{app:label_space}

This section describes the label taxonomy used in our relation-aware ESC benchmark.
Our goal is not to reproduce a full clinical coding system, but to construct a compact task-oriented taxonomy for evaluating whether models can recognize relational dynamics and make relation-sensitive support decisions.
We therefore organize the taxonomy around two clinical anchors: dyadic couple processes and systemic family processes.
The taxonomy contains two core dimensions: \textit{relation patterns}, which describe the dominant interpersonal dynamics at a given moment, and \textit{support strategies}, which describe the therapist's primary intervention intention in therapist turns.

\paragraph{Theoretical grounding.}
For couple conversations, the primary theoretical anchor is Emotionally Focused Therapy (EFT), which conceptualizes couple distress as recurring negative interaction cycles and emphasizes emotional access, softening, enactment, and relational repair \citep{johnson2012practice,johnson2006path,bradley2004toward}.
We use EFT to define central couple-side relation patterns such as \texttt{pursue\_withdraw}, \texttt{attack\_attack}, and \texttt{repair\_softening}, as well as support strategies such as \texttt{track}, \texttt{evoke}, \texttt{enact}, and \texttt{repair}.
Integrative Behavioral Couple Therapy (IBCT) is used as a secondary anchor for labels that require partners to jointly observe their interaction pattern, such as \texttt{detach} and \texttt{constructive\_alignment} \citep{christensen1995integrative,jacobson1996integrative}.
Gottman-style conflict-repair research further supports labels related to escalation, mutual attack, withdrawal, and repair \citep{gottman2023predicts,gottman2016marriage}.
Together, these sources provide a coherent couple-side taxonomy centered on negative cycles, emotional softening, shared pattern awareness, and relational repair.

For family conversations, the primary theoretical anchor is Structural Family Therapy (SFT), which explains family difficulties through boundaries, hierarchy, subsystems, coalitions, and in-session interactional organization \citep{minuchin2018families,minuchin1981family,nichols2013techniques}.
We use SFT to define family-specific relation patterns such as \texttt{boundary\_hierarchy\_strain}, \texttt{cross\_generational\_coalition}, and \texttt{mutual\_disengagement}, as well as support strategies such as \texttt{join}, \texttt{enact}, \texttt{boundary}, and \texttt{counterbalance}.
Family Systems Theory (FST) provides a complementary framework for representing recurring family-level interaction cycles, coalitions, disengagement, and systemic reframing \citep{bowen1993family,cox1997families}.
EFT is additionally used for emotion-focused labels involving pursuit-withdrawal, vulnerability, softening, and repair.
Thus, the family-side taxonomy is organized around family structure, systemic interaction patterns, and relational reorganization rather than isolated individual symptoms.

\paragraph{Construction principles.}
We construct the label space according to three principles.
First, each label should correspond to a clinically meaningful relational construct rather than a surface-level conversational phenomenon.
For example, \texttt{pursue\_withdraw} captures a recurrent cycle of pursuit and avoidance, not merely a sequence in which one speaker talks more than another.
Second, labels should be operationally distinguishable in observable dialogue and multimodal evidence.
For instance, \texttt{repair\_softening} denotes an emerging movement toward vulnerability or reconnection, whereas \texttt{constructive\_alignment} denotes a more stable shared stance toward the problem.
Third, the taxonomy should remain compact enough for reliable annotation and benchmark evaluation.
Therefore, fine-grained clinical concepts are grouped into a smaller number of task-oriented categories.

\begin{table*}[t]
\centering
\small
\setlength{\tabcolsep}{4pt}
\renewcommand{\arraystretch}{1.12}
\resizebox{0.9\linewidth}{!}{
\begin{tabular}{p{3.4cm}p{7.4cm}p{2cm}}
\toprule
\textbf{Label} & \textbf{Operational Meaning} & \textbf{Main Basis} \\
\midrule
\texttt{pursue\_withdraw}
& One partner presses, demands, pursues, or seeks engagement, while the other avoids, shuts down, minimizes, or disengages.
& EFT \\

\texttt{attack\_attack}
& Both partners criticize, blame, defend, or counterattack, leading to mutual escalation.
& EFT; Gottman \\

\texttt{withdraw\_withdraw}
& Both partners avoid emotional engagement, show low responsiveness, or mutually disengage from the issue.
& EFT; Gottman \\

\texttt{repair\_softening}
& Defensiveness decreases and the interaction begins to show vulnerability, apology, validation, emotional openness, or reconnection.
& EFT; Gottman \\

\texttt{constructive\_alignment}
& Partners move toward shared understanding, cooperation, or joint problem solving.
& EFT; IBCT \\

\texttt{mixed\_transition}
& Multiple relational signals coexist, or the interaction is clearly shifting between two relation patterns.
& EFT; IBCT \\
\bottomrule
\end{tabular}}
\caption{Couple relation pattern labels and their main theoretical grounding. EFT = Emotionally Focused Therapy; Gottman = Gottman Method and conflict-repair research; IBCT = Integrative Behavioral Couple Therapy.}
\label{tab:couple_relation_pattern_label_space}
\end{table*}
\begin{table*}[t]
\centering
\small
\setlength{\tabcolsep}{4pt}
\renewcommand{\arraystretch}{1.12}
\resizebox{0.9\linewidth}{!}{
\begin{tabular}{p{4.4cm}p{7.2cm}p{2cm}}
\toprule
\textbf{Label} & \textbf{Operational Meaning} & \textbf{Main Basis} \\
\midrule
\texttt{escalation\_conflict}
& Two or more family members engage in overt conflict, argument, blame, or escalating emotional exchange.
& SFT; FST \\

\texttt{pursue\_withdraw}
& One family member presses, demands, pursues, or seeks engagement, while another avoids, shuts down, or disengages.
& FST; EFT \\

\texttt{mutual\_disengagement}
& Key family members collectively withdraw, remain silent, avoid the central issue, or show low participation.
& SFT; FST \\

\texttt{cross\_generational\_coalition}
& Members across generations form an alliance around or against another family member.
& SFT; FST \\

\texttt{boundary\_hierarchy\_strain}
& The interaction is organized by unclear boundaries, unstable hierarchy, role reversal, subsystem tension, or parent-child role confusion.
& SFT \\

\texttt{repair\_softening}
& Defensiveness, blame, or rigidity decreases, and the family interaction begins to show vulnerability, validation, apology, or reconnection.
& EFT; FST \\

\texttt{cooperative\_family\_alliance}
& Multiple family members form a coordinated, constructive, and shared stance toward the problem.
& FST \\

\texttt{mixed\_transition}
& Multiple family dynamics are simultaneously salient, or the interaction is clearly shifting between relation patterns.
& SFT; FST \\
\bottomrule
\end{tabular}}
\caption{Family relation pattern labels and their main theoretical grounding. SFT = Structural Family Therapy; FST = Family Systems Theory; EFT = Emotionally Focused Therapy.}
\label{tab:family_relation_pattern_label_space}
\end{table*}

\subsection{Relation Pattern Labels}
\label{app:relation_pattern_labels}

Relation-pattern labels describe the dominant relational cycle or interpersonal configuration at a target moment.
Because couple and family conversations differ in their relational structure, we define separate but partially overlapping label sets for the two scenarios.
Tables~\ref{tab:couple_relation_pattern_label_space} and~\ref{tab:family_relation_pattern_label_space} summarize the relation-pattern labels for couple and family scenarios, respectively.

\paragraph{Couple relation patterns.}
Couple relation-pattern labels mainly capture dyadic negative cycles, withdrawal patterns, repair attempts, and constructive coordination. Based on this, we summarize and define six representative patterns as shown in Table~\ref{tab:couple_relation_pattern_label_space}, detailed as follows.

\texttt{pursue\_withdraw} refers to a negative cycle in which one partner seeks engagement, explanation, or emotional response, while the other avoids, minimizes, shifts topics, or shuts down. The label is used when the interaction is mainly organized by this asymmetric pattern of pressure and retreat.

\texttt{attack\_attack} refers to a mutually escalating conflict cycle in which both partners respond to each other through blame, criticism, defensiveness, or counterattack. Unlike \texttt{pursue\_withdraw}, where one side tends to disengage, both partners remain actively involved in the conflict and the interaction is organized around reciprocal attack.

\texttt{withdraw\_withdraw} describes a mutually disengaged pattern in which both partners avoid emotional contact or show little willingness to enter the central issue. This pattern is often marked by silence, minimal responses, topic avoidance, emotional flatness, or parallel withdrawal, and differs from \texttt{pursue\_withdraw} because neither partner is actively pursuing engagement.

\texttt{repair\_softening} captures moments when a negative cycle begins to loosen. Typical evidence includes reduced defensiveness, softer tone, expressions of vulnerability, apology, validation, acknowledgment of hurt, or willingness to reconnect. This label is used for emerging repair attempts rather than fully established cooperation.

\texttt{constructive\_alignment} refers to a more stable cooperative stance in which partners begin to understand the problem as shared or jointly manageable. Compared with \texttt{repair\_softening}, it indicates that the interaction has moved beyond an initial softening moment toward clearer collaboration, perspective-taking, or joint problem solving.

\texttt{mixed\_transition} is used when the segment contains competing relational signals or a clear shift between patterns. For example, a partner may show vulnerability while the other remains defensive, or the conversation may move from escalation into partial repair without forming a stable new pattern. This label captures transitional moments that cannot be faithfully represented by a single dominant cycle.

\begin{table*}[t]
\centering
\small
\setlength{\tabcolsep}{4pt}
\renewcommand{\arraystretch}{1.12}
\resizebox{0.85\linewidth}{!}{
\begin{tabular}{p{3cm}p{7.5cm}p{2cm}}
\toprule
\textbf{Label} & \textbf{Operational Meaning} & \textbf{Main Basis} \\
\midrule
\texttt{track}
& The therapist follows, names, or clarifies the ongoing interactional process or relational cycle.
& EFT \\

\texttt{reframe}
& The therapist changes the meaning of an event or interaction, often shifting from individual blame to a shared relational pattern.
& EFT; IBCT \\

\texttt{evoke}
& The therapist deepens access to primary emotions, attachment needs, vulnerability, shame, fear, or longing.
& EFT \\

\texttt{enact}
& The therapist invites one partner to speak directly to the other rather than only speaking to the therapist.
& EFT \\

\texttt{join}
& The therapist promotes empathic connection, emotional contact, or working alliance between partners.
& EFT; IBCT \\

\texttt{detach}
& The therapist helps partners step back and jointly observe the problem as a shared interaction pattern rather than treating each other as the problem.
& IBCT \\

\texttt{repair}
& The therapist supports apology, acknowledgment of hurt, clarification of intention, emotional repair, or reconnection.
& EFT\\

\texttt{counterbalance}
& The therapist restores voice, space, or participation for a less powerful, less heard, or interactionally disadvantaged partner.
& WAT \\

\texttt{safeguard}
& The therapist protects a vulnerable partner from being shamed, attacked, overwhelmed, or emotionally flooded.
& WAT \\

\texttt{goal\_align}
& The therapist redirects partners toward a shared therapeutic task, common goal, or collaborative working frame.
& WAT \\
\bottomrule
\end{tabular}}
\caption{Couple support strategy labels and their main theoretical grounding. EFT = Emotionally Focused Therapy; IBCT = Integrative Behavioral Couple Therapy; WAT = Working Alliance Theory.}
\label{tab:couple_support_strategy_label_space}
\end{table*}

\paragraph{Family relation patterns.}
Family relation-pattern labels extend the dyadic view to multi-party systemic configurations. Based on this, we summarize and define eight representative patterns as shown in Table~\ref{tab:family_relation_pattern_label_space}, detailed as follows.

\texttt{escalation\_conflict} refers to an overt family conflict pattern in which two or more members intensify disagreement through blame, interruption, criticism, defensiveness, or emotional confrontation. The label is used when the interaction is primarily organized by rising conflict intensity rather than by withdrawal or repair.

\texttt{pursue\_withdraw} describes an asymmetric family interaction in which one member seeks engagement, explanation, or response, while another avoids, minimizes, remains silent, or disengages. It is retained from the couple taxonomy because similar pressure--retreat cycles can also organize parent-child or other family interactions.

\texttt{mutual\_disengagement} captures a family-level withdrawal pattern in which key members collectively avoid the central issue or show low emotional participation. Typical evidence includes silence, brief responses, topic avoidance, emotional flatness, or a general lack of willingness to engage with one another.

\texttt{cross\_generational\_coalition} refers to a systemic configuration in which members from different generations align with each other in a way that marginalizes, opposes, or places pressure on another member. The label is used when the interaction is shaped by coalition or triangulation rather than by a simple dyadic conflict.

\texttt{boundary\_hierarchy\_strain} describes family interactions organized by unclear boundaries, unstable hierarchy, role confusion, or subsystem tension. Typical cases include parent-child role reversal, a child being pulled into adult conflict, or family members crossing generational or relational boundaries.

\texttt{repair\_softening} marks moments when a rigid or conflictual family interaction begins to loosen. Evidence may include reduced blame, softer tone, apology, validation, acknowledgment of hurt, vulnerability, or an expressed willingness to reconnect, but the repair is still emerging rather than fully stabilized.

\texttt{cooperative\_family\_alliance} refers to a constructive family-level stance in which multiple members begin to coordinate around shared understanding, mutual support, or joint problem solving. Compared with \texttt{repair\_softening}, this label indicates a more stable cooperative orientation across the family system.

\texttt{mixed\_transition} is used when multiple family dynamics are simultaneously salient or when the interaction is clearly shifting between patterns. For example, a segment may contain both coalition and repair signals, or move from escalation toward partial cooperation without a single stable dominant pattern.

\begin{table*}[t]
\centering
\small
\setlength{\tabcolsep}{4pt}
\renewcommand{\arraystretch}{1.12}
\resizebox{0.85\linewidth}{!}{
\begin{tabular}{p{2.7cm}p{7.5cm}p{2cm}}
\toprule
\textbf{Label} & \textbf{Operational Meaning} & \textbf{Main Basis} \\
\midrule
\texttt{join}
& The therapist builds trust, affiliation, and working alliance with the family system.
& SFT \\

\texttt{track}
& The therapist follows and names recurring family processes, such as escalation, alliance formation, disengagement, or triangulation.
& SFT; FST \\

\texttt{counterbalance}
& The therapist brings a less-heard, lower-power, or structurally marginalized family member back into the interaction.
& SFT \\

\texttt{enact}
& The therapist invites family members to interact directly in session so that the relational pattern can be observed and reorganized.
& SFT \\

\texttt{boundary}
& The therapist clarifies or reorganizes boundaries, hierarchy, roles, subsystems, or intergenerational structure.
& SFT \\

\texttt{reframe}
& The therapist shifts the meaning of a problem from individual blame to a systemic or shared family process.
& FST; SFT \\

\texttt{repair}
& The therapist supports apology, recognition, emotional repair, or reconnection among family members.
& EFT; FST \\

\texttt{safeguard}
& The therapist protects a vulnerable family member from being scapegoated, attacked, shamed, or emotionally overwhelmed.
& SFT \\
\bottomrule
\end{tabular}}
\caption{Family support strategy labels and their main theoretical grounding. SFT = Structural Family Therapy; FST = Family Systems Theory; EFT = Emotionally Focused Therapy.}
\label{tab:family_support_strategy_label_space}
\end{table*}

\subsection{Support Strategy Labels}
\label{app:support_strategy_labels}

Support strategy labels are annotated only for therapist turns and describe the therapist's primary intervention intention.
Because couple and family therapy emphasize different relational organizations, we use separate closed-set strategy labels for couple and family scenarios.
Tables~\ref{tab:couple_support_strategy_label_space} and~\ref{tab:family_support_strategy_label_space} summarize the support-strategy labels for couple and family scenarios, respectively.

\paragraph{Couple support strategies.}
The couple support-strategy labels are organized around EFT-style cycle work, with IBCT and Gottman-style repair research serving as focused supplements. As shown in Table~\ref{tab:couple_support_strategy_label_space}, we summarize and define 10 representative strategies in total. 

\texttt{track} refers to interventions that follow, name, or clarify the ongoing interactional cycle between partners. The therapist uses this strategy to make the relational process visible, such as identifying how one partner's pursuit and the other's withdrawal reinforce each other.

\texttt{reframe} refers to interventions that change the meaning of an event, emotion, or interaction. The therapist often shifts the focus from individual blame to a shared relation pattern, helping partners see the conflict as something they are caught in together rather than as one person's fault.

\texttt{evoke} refers to interventions that deepen access to primary emotions and attachment needs. The therapist encourages a partner to move beyond surface anger or defensiveness and articulate more vulnerable feelings, such as hurt, fear, shame, loneliness, or longing for connection.

\texttt{enact} refers to interventions that invite one partner to speak directly to the other in session. Instead of talking only to the therapist, the speaker is guided to express an emotion, need, request, or acknowledgment directly to their partner.

\texttt{join} refers to interventions that promote empathic connection and emotional contact between partners. The therapist supports moments in which partners can understand, receive, or respond to each other's emotional experience in a softer and more engaged way.

\texttt{detach} refers to interventions that help partners step back from immediate blame or reactivity and jointly observe their interaction pattern. The therapist frames the problem as a shared cycle that both partners can examine, rather than as a defect or failure of one partner.

\texttt{repair} refers to interventions that guide partners toward apology, clarification, acknowledgment of hurt, or relational reconnection. The therapist uses this strategy when the interaction calls for restoring trust, addressing injury, or helping partners respond constructively after conflict.

\texttt{counterbalance} refers to interventions that restore voice, space, or participation for a less-heard or interactionally disadvantaged partner. The therapist uses this strategy to rebalance the conversation when one partner dominates, dismisses, interrupts, or leaves little room for the other to express their experience.

\texttt{safeguard} refers to interventions that protect a vulnerable partner from being overwhelmed, shamed, attacked, or emotionally flooded. The therapist may slow down the interaction, interrupt harmful exchanges, or create enough safety for the partner to remain engaged.

\texttt{goal\_align} refers to interventions that redirect partners toward a shared therapeutic task, common goal, or collaborative working frame. The therapist uses this strategy when the conversation drifts into blame, defensiveness, or side conflicts and needs to be reoriented toward joint relational work.

\paragraph{Family support strategies.}
The family support-strategy labels are organized primarily around SFT, with FST providing a complementary systemic view of recurring family interaction patterns. As shown in Table~\ref{tab:family_support_strategy_label_space}, we summarize 8 representative strategies and their detailed definition are listed below.

\texttt{join} refers to interventions that build contact, trust, and working engagement with the family system. The therapist uses this strategy to enter the family interaction in a supportive way and establish enough connection for members to participate in the therapeutic work.

\texttt{track} refers to interventions that follow, name, or clarify recurring family processes. The therapist may identify patterns such as escalation, disengagement, triangulation, coalition formation, or repeated parent-child conflict so that the family can recognize the interactional process organizing the problem.

\texttt{enact} refers to interventions that invite family members to interact directly with each other in session. This allows the therapist to observe the live relation pattern and guide members toward a different way of responding.

\texttt{boundary} refers to structural interventions that clarify or reorganize boundaries, hierarchy, roles, subsystems, or intergenerational organization. The therapist uses this strategy when the problem involves role confusion, parent-child boundary issues, inappropriate coalition, or unclear family structure.

\texttt{reframe} refers to interventions that shift the meaning of a problem from individual blame to a systemic or shared family process. The therapist helps members see a behavior or conflict as part of a broader interactional pattern rather than as the fault of one person.

\texttt{repair} refers to interventions that support apology, recognition, emotional repair, or reconnection among family members. The therapist uses this strategy when the family interaction shows an opportunity to acknowledge hurt, reduce defensiveness, or rebuild trust.

\texttt{safeguard} refers to interventions that protect a vulnerable family member from being scapegoated, attacked, shamed, or emotionally overwhelmed. The therapist may slow down the exchange, interrupt harmful interaction, or create safety for the member to stay engaged.

\texttt{counterbalance} refers to interventions that bring a less-heard, lower-power, or structurally marginalized family member back into the interaction. The therapist uses this strategy to rebalance participation when one person is dominated, ignored, or excluded from the family conversation.

\subsection{Label Boundaries and Disambiguation Rules}
\label{app:label_boundaries}

This subsection clarifies the boundaries between conceptually adjacent labels.
Because several labels may share similar surface cues, the label decision is based on the central relational function of the segment: for relation patterns, what interactional organization dominates the moment; for support strategies, what primary intervention function the therapist turn performs.
Surface wording alone is not sufficient for label assignment.

\paragraph{Boundaries between support-strategy labels.}
\texttt{track} and \texttt{reframe} both involve describing an interactional process, but they differ in whether the therapist changes its meaning.
A therapist turn is labeled as \texttt{track} when it maps, names, or follows a live sequence or recurring cycle without substantially altering its interpretation.
It is labeled as \texttt{reframe} when the therapist transforms the same sequence from individual blame into a shared relational, attachment-based, or systemic formulation.

\texttt{evoke} and \texttt{join} are both emotion-oriented strategies, but they target different relational functions.
\texttt{evoke} is used when the therapist deepens one participant's primary emotion, such as hurt, fear, shame, or longing.
\texttt{join} is used when the therapist helps that emotion become hearable, receivable, or connective within the couple or family relationship.

In family sessions, \texttt{track} and \texttt{boundary} are distinguished by whether the therapist only identifies a pattern or actively reorganizes the family structure.
A turn remains \texttt{track} when it describes escalation, disengagement, triangulation, or coalition.
It is labeled as \texttt{boundary} when the therapist intervenes in roles, generational boundaries, subsystem relations, hierarchy, or parent-child organization.

\texttt{counterbalance} and \texttt{safeguard} both address asymmetry in the interaction, but they differ in urgency and function.
\texttt{counterbalance} is used when the therapist restores voice, access, or influence to a less-heard or lower-power participant.
\texttt{safeguard} is reserved for moments where the primary function is immediate protection from emotional flooding, humiliation, scapegoating, attack, or escalation.

\paragraph{Boundaries between relation-pattern labels.}
\texttt{repair\_softening}, \texttt{constructive\_alignment}, and \texttt{cooperative\_family\_alliance} all indicate movement away from negative cycles, but they represent different degrees of stabilization.
\texttt{repair\_softening} marks an early and still-fragile movement out of blame, shutdown, rupture, or disconnection.
By contrast, \texttt{constructive\_alignment} in couple sessions and \texttt{cooperative\_family\_alliance} in family sessions require a more sustained shared stance toward the problem.
A mere pause in conflict is not sufficient for these positive-state labels if hostility, withdrawal, coalition, or structural strain still organizes the segment.

\texttt{attack\_attack} and \texttt{pursue\_withdraw} are separated by whether both sides actively escalate.
\texttt{attack\_attack} is used when both partners are engaged in reciprocal blame, criticism, defensiveness, or counterattack.
\texttt{pursue\_withdraw} is used when one participant presses for engagement while the other avoids, minimizes, shuts down, or retreats.
In family sessions, broad reciprocal escalation involving multiple members is labeled as \texttt{escalation\_conflict} rather than \texttt{pursue\_withdraw}.

\texttt{cross\_generational\_coalition} and \texttt{boundary\_hierarchy\_strain} both involve family structure, but they capture different configurations.
\texttt{cross\_generational\_coalition} is used when a specific cross-generational alliance or triangle organizes the interaction.
\texttt{boundary\_hierarchy\_strain} is used when the sharper issue is broader role confusion, parentification, unstable hierarchy, or unclear subsystem boundaries without one dominant coalition.

Finally, \texttt{mixed\_transition} is not a fallback label for uncertainty or insufficient evidence.
It is used only when two strong patterns are simultaneously salient, or when the segment falls at a clinically visible pivot point between patterns.
If one pattern clearly organizes the interaction, that pattern remains the primary label.

\section{Scenario-based Detailed Results}
\label{app:scenario_based_results}

\subsection{Scenario-wise Performance}
Table~\ref{tab:appendix-scenario-wise-performance} shows that scenario differences are mixed rather than uniform across tasks. At the average level, couple and family conversations are very close on ITP (82.60\% vs.\ 82.70\% F1), while family conversations are slightly higher on RPP (41.23\% vs.\ 40.20\% accuracy). By contrast, the support-side metrics shift in different directions across scenarios: STP is lower in family conversations than in couple conversations (54.87\% vs.\ 66.88\% recall; 69.28\% vs.\ 80.79\% MRR), whereas SSP is higher in family conversations on average (36.83\% vs.\ 30.54\% recall; 50.56\% vs.\ 43.45\% MRR). ER remains nearly unchanged across scenarios (4.03 vs.\ 4.05), while VP is slightly higher in family conversations (3.76 vs.\ 3.56).

The scenario effect also varies substantially by model. For RPP, some models drop from couple to family conversations, such as \texttt{Qwen-Plus} (40.58\% to 29.98\%) and \texttt{GPT-4o} (41.46\% to 28.75\%), while others improve, including \texttt{DeepSeek-V4-Flash} (36.56\% to 49.56\%), \texttt{Kimi K2.5} (44.11\% to 49.38\%), and \texttt{MiniMax M2.5} (38.02\% to 45.15\%). A similar inconsistency appears in SSP: \texttt{GPT-4o} achieves the best recall in couple conversations (36.28\%), whereas \texttt{DeepSeek-V4-Pro} performs best in family conversations (42.65\%). These results indicate that scenario differences do not produce a single consistent ranking across models, but instead interact with the specific relational cues captured by each model.

\begin{table*}[t]
\centering
\small
\setlength{\tabcolsep}{3.0pt}
\begin{tabular}{llccc c cc cc cc cc}
\toprule
\multirow{2}{*}{Model}
& \multirow{2}{*}{Scenario}
& \multicolumn{3}{c}{ITP}
& \multicolumn{1}{c}{RPP}
& \multicolumn{2}{c}{STP}
& \multicolumn{2}{c}{SSP}
& \multicolumn{2}{c}{ER}
& \multicolumn{2}{c}{VP} \\
\cmidrule(lr){3-5}
\cmidrule(lr){6-6}
\cmidrule(lr){7-8}
\cmidrule(lr){9-10}
\cmidrule(lr){11-12}
\cmidrule(lr){13-14}
& & Prec. & Rec. & F1 & Acc. & Rec. & MRR & Rec. & MRR & LLM & BERT & LLM & BERT \\
\midrule
\texttt{Qwen-Plus} & Couple & 55.95 & 90.72 & 69.22 & 40.58 & 68.45 & 81.81 & 23.49 & 33.81 & 3.54 & 0.8155 & 3.18 & 0.8599 \\
\texttt{Qwen-Plus} & Family & 53.66 & 81.48 & 64.71 & 29.98 & 55.04 & 69.55 & 36.31 & 48.61 & 4.04 & 0.8289 & 3.41 & 0.8471 \\
\texttt{Qwen-Plus} & Both & 55.78 & 89.99 & 68.87 & 38.27 & 65.61 & 79.21 & 26.21 & 36.95 & 3.58 & 0.8167 & 3.20 & 0.8587 \\
\midrule
\texttt{Qwen3-Max} & Couple & 80.30 & 70.39 & 75.02 & 43.23 & 63.64 & 78.77 & 30.08 & 44.11 & 4.05 & 0.8219 & 3.30 & 0.8659 \\
\texttt{Qwen3-Max} & Family & 75.69 & 76.76 & 76.22 & 43.21 & 51.59 & 67.00 & 38.62 & 51.92 & 3.94 & 0.8278 & 3.80 & 0.8516 \\
\texttt{Qwen3-Max} & Both & 79.81 & 70.99 & 75.14 & 43.22 & 61.09 & 76.28 & 31.89 & 45.76 & 4.04 & 0.8224 & 3.35 & 0.8646 \\
\midrule
\texttt{Qwen3.5-Plus} & Couple & 97.88 & 91.48 & 94.57 & 39.25 & 72.09 & 84.07 & 31.16 & 44.53 & 4.22 & 0.8316 & 3.95 & 0.8656 \\
\texttt{Qwen3.5-Plus} & Family & 99.23 & 90.85 & 94.85 & 43.92 & 59.37 & 72.29 & 33.72 & 46.64 & 4.16 & 0.8373 & 3.84 & 0.8494 \\
\texttt{Qwen3.5-Plus} & Both & 98.00 & 91.42 & 94.60 & 40.27 & 69.40 & 81.57 & 31.70 & 44.98 & 4.22 & 0.8321 & 3.94 & 0.8641 \\
\midrule
\texttt{DeepSeek-R1} & Couple & 90.85 & 72.23 & 80.47 & 37.98 & 63.72 & 78.91 & 28.68 & 38.55 & 4.09 & 0.8221 & 3.31 & 0.8648 \\
\texttt{DeepSeek-R1} & Family & 89.38 & 71.13 & 79.22 & 37.57 & 55.04 & 69.31 & 31.99 & 44.00 & 4.04 & 0.8295 & 3.71 & 0.8500 \\
\texttt{DeepSeek-R1} & Both & 90.71 & 72.12 & 80.36 & 37.89 & 61.88 & 76.88 & 29.38 & 39.71 & 4.08 & 0.8227 & 3.35 & 0.8634 \\
\midrule
\texttt{DeepSeek-V3.2} & Couple & 93.64 & 81.19 & 86.97 & 37.00 & 65.04 & 79.56 & 31.40 & 44.08 & 4.14 & 0.8257 & 3.86 & 0.8599 \\
\texttt{DeepSeek-V3.2} & Family & 91.27 & 80.99 & 85.82 & 32.80 & 53.89 & 68.16 & 39.19 & 53.94 & 4.10 & 0.8287 & 3.92 & 0.8482 \\
\texttt{DeepSeek-V3.2} & Both & 93.42 & 81.17 & 86.86 & 36.08 & 62.68 & 77.14 & 33.05 & 46.17 & 4.13 & 0.8259 & 3.86 & 0.8588 \\
\midrule
\texttt{DeepSeek-V4-Flash} & Couple & 90.85 & 78.10 & 84.00 & 36.56 & 65.66 & 80.37 & 28.22 & 42.17 & 4.05 & 0.8151 & 4.12 & 0.8587 \\
\texttt{DeepSeek-V4-Flash} & Family & 93.44 & 80.28 & 86.36 & 49.56 & 51.30 & 67.34 & 37.18 & 51.39 & 3.91 & 0.8169 & 3.73 & 0.8478 \\
\texttt{DeepSeek-V4-Flash} & Both & 91.10 & 78.31 & 84.22 & 39.39 & 62.61 & 77.61 & 30.12 & 44.13 & 4.04 & 0.815 & 4.08 & 0.8576 \\
\midrule
\texttt{DeepSeek-V4-Pro} & Couple & 92.59 & 75.24 & 83.02 & 43.82 & 73.41 & 84.46 & 35.81 & 50.84 & 4.18 & 0.8238 & 3.88 & 0.8582 \\
\texttt{DeepSeek-V4-Pro} & Family & 89.31 & 82.39 & 85.71 & 52.03 & 62.25 & 73.97 & 42.65 & 56.29 & 4.07 & 0.8324 & 4.01 & 0.8485 \\
\texttt{DeepSeek-V4-Pro} & Both   & 92.24 & 75.91 & 83.28 & 45.60 & 71.04 & 82.23 & 37.26 & 52.00 & 4.17 & 0.8246 & 3.90 & 0.8573 \\
\midrule
\texttt{Kimi K2.5} & Couple & 88.32 & 83.32 & 85.75 & 44.11 & 61.16 & 77.75 & 33.33 & 46.51 & 4.18 & 0.8222 & 3.62 & 0.8638 \\
\texttt{Kimi K2.5} & Family & 85.43 & 90.85 & 88.05 & 49.38 & 51.59 & 67.58 & 37.18 & 51.92 & 4.13 & 0.8259 & 3.81 & 0.8443 \\
\texttt{Kimi K2.5} & Both & 88.01 & 84.03 & 85.98 & 45.26 & 59.13 & 75.60 & 34.15 & 47.66 & 4.18 & 0.8225 & 3.64 & 0.8619 \\
\midrule
\texttt{MiniMax M2.5} & Couple & 90.12 & 75.75 & 82.32 & 38.02 & 65.35 & 79.70 & 26.98 & 40.76 & 4.01 & 0.8195 & 3.21 & 0.8574 \\
\texttt{MiniMax M2.5} & Family & 85.38 & 78.17 & 81.62 & 45.15 & 52.16 & 67.00 & 36.89 & 50.82 & 3.98 & 0.8256 & 3.59 & 0.8428 \\
\texttt{MiniMax M2.5} & Both & 89.64 & 75.98 & 82.25 & 39.85 & 62.55 & 77.01 & 29.08 & 42.89 & 4.01 & 0.8201 & 3.25 & 0.8561 \\
\midrule
\texttt{GPT-4o} & Couple & 87.09 & 82.29 & 84.62 & 41.46 & 70.31 & 82.52 & 36.28 & 49.17 & 4.02 & 0.8200 & 3.19 & 0.8684 \\
\texttt{GPT-4o} & Family & 82.99 & 85.92 & 84.43 & 28.75 & 56.48 & 70.65 & 34.58 & 50.05 & 3.90 & 0.8258 & 3.74 & 0.8573 \\
\texttt{GPT-4o} & Both & 86.67 & 82.63 & 84.60 & 38.69 & 67.38 & 80.00 & 35.92 & 49.36 & 4.01 & 0.8206 & 3.24 & 0.8674 \\
\midrule
Average & Couple & 86.76 & 80.07 & 82.60 & 40.20 & 66.88 & 80.79 & 30.54 & 43.45 & 4.05 & 0.8217 & 3.56 & 0.8623 \\
Average & Family & 84.58 & 81.88 & 82.70 & 41.23 & 54.87 & 69.28 & 36.83 & 50.56 & 4.03 & 0.8279 & 3.76 & 0.8487 \\
Average & Both   & 86.54 & 80.25 & 82.62 & 40.45 & 64.34 & 78.35 & 31.88 & 44.96 & 4.05 & 0.8223 & 3.58 & 0.8610 \\
\bottomrule
\end{tabular}
\caption{Scenario-wise performance across relation-aware ESC tasks.}
\label{tab:appendix-scenario-wise-performance}
\end{table*}

\subsection{Scenario-based Label Distribution}

\begin{figure}[t]
  \centering
  \includegraphics[width=\linewidth]{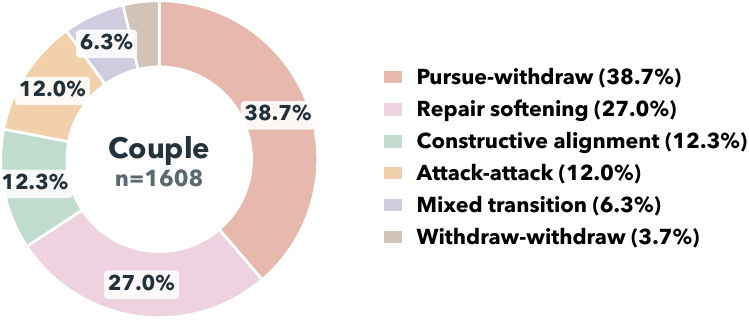}
  \caption{Distribution of relation-pattern labels in couple conversations. Couple interactions are dominated by \textit{pursue-withdraw} and \textit{repair softening}, with smaller shares of \textit{constructive alignment}, \textit{attack-attack}, and transitional states.}
  \label{fig:relation_pattern_distribution_couple}
\end{figure}

\begin{figure}[t]
  \centering
  \includegraphics[width=\linewidth]{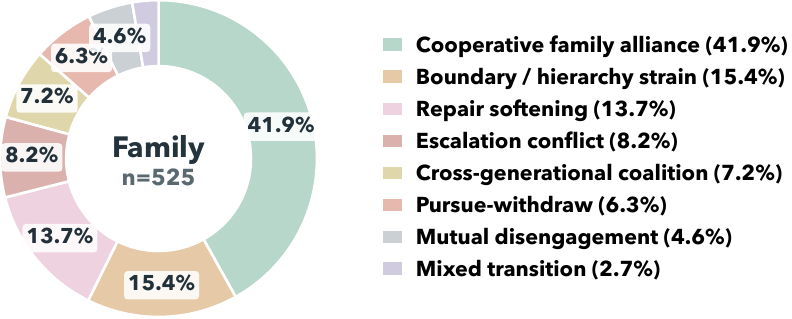}
  \caption{Distribution of relation-pattern labels in family conversations. Family interactions are dominated by \textit{cooperative family alliance}, followed by \textit{boundary / hierarchy strain} and \textit{repair softening}, reflecting a different relational structure from couple conversations.}
  \label{fig:relation_pattern_distribution_family}
\end{figure}

\begin{figure}[t]
  \centering
  \includegraphics[width=\linewidth]{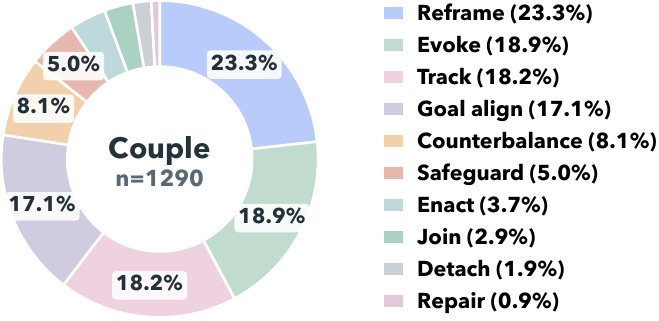}
  \caption{Distribution of support-strategy labels in couple conversations. Therapist responses in couple sessions are concentrated in \textit{reframe}, \textit{evoke}, \textit{track}, and \textit{goal align}, while other strategies appear much less frequently.}
  \label{fig:support_strategy_distribution_couple}
\end{figure}

\begin{figure}[t]
  \centering
  \includegraphics[width=\linewidth]{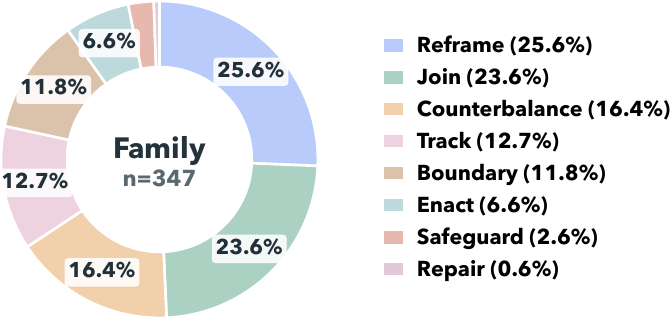}
  \caption{Distribution of support-strategy labels in family conversations. Family sessions show a broader emphasis on \textit{reframe}, \textit{join}, \textit{counterbalance}, \textit{track}, and \textit{boundary}, indicating a different intervention profile from couple conversations.}
  \label{fig:support_strategy_distribution_family}
\end{figure}

Figures~\ref{fig:relation_pattern_distribution_couple}--\ref{fig:support_strategy_distribution_family} reveal a clear long-tail distribution in both scenarios. In couple conversations, relation-pattern labels are highly concentrated in \textit{pursue-withdraw} (38.7\%) and \textit{repair softening} (27.0\%), while the two rarest states, \textit{mixed transition} and \textit{withdraw-withdraw}, together account for less than 10\% of the data. Family conversations show a similar imbalance: \textit{cooperative family alliance} alone accounts for 41.9\%, while five labels remain below 10\%. The support-strategy labels are also strongly skewed. In couple data, the four most frequent strategies (\textit{reframe}, \textit{evoke}, \textit{track}, and \textit{goal align}) cover 77.4\% of all strategy instances, and in family data the five most frequent strategies (\textit{reframe}, \textit{join}, \textit{counterbalance}, \textit{track}, and \textit{boundary}) cover 90.2\%. By contrast, fine-grained strategies such as \textit{repair}, \textit{enact}, and \textit{safeguard} appear only rarely.

This label imbalance is reflected in model behavior. The per-label RPP results show that dominant and relatively salient relation patterns are generally much easier to predict than rare and fine-grained ones. In couple conversations, the most frequent pattern, \textit{pursue-withdraw}, reaches 77.38\% average accuracy, whereas the much rarer \textit{mixed transition} reaches only 2.23\%. In family conversations, \textit{pursue-withdraw} (72.73\%) and \textit{cooperative family alliance} (54.24\%) are recognized much more reliably than \textit{escalation conflict} (14.73\%), \textit{cross-generational coalition} (17.25\%), \textit{mutual disengagement} (14.81\%), and \textit{mixed transition} (1.59\%). This suggests that models are better at recovering common, recurrent relational configurations, but struggle with sparse labels that require finer discrimination between subtle or transitional group states. One notable exception is \textit{repair softening}: although it is frequent in both scenarios, it remains only moderately predictable (35.45\% in couple and 48.46\% in family), indicating that semantic overlap between nearby de-escalatory states also contributes to errors. The same long-tail effect likely contributes to the weak SSP results, where models appear more capable of defaulting to common strategy families than of reliably identifying sparse, fine-grained intervention types.

\section{Human Evaluation For LLM-as-judge}
\label{app:human_eval}
\begin{table*}[t]
\centering
\caption{Human Evaluation Results}
\label{tab:human-evaluation}
\begin{tabular}{lrrrrrrrr}
\hline
Annotator & Task & \# Samples & Exact Match & Diff = 1 & Diff = 2 & Diff = 3 & Diff = 4 & MAE \\
\hline
Annotator1 & VP & 50 & 39 & 10 & 1 & 0 & 0 & 0.24 \\
Annotator2   & VP & 50 & 42 & 8  & 0 & 0 & 0 & 0.16 \\
Annotator3   & ER & 50 & 38 & 9  & 3 & 0 & 0 & 0.30 \\
Annotator4   & ER & 50 & 40 & 8  & 2 & 0 & 0 & 0.24 \\
\hline
Total        & VP,ER &200 & 159 & 35 & 6 & 0 & 0 & 0.235 \\
\hline
\end{tabular}
\end{table*}

\begin{figure*}[t]
    \centering
    \includegraphics[width=\linewidth]{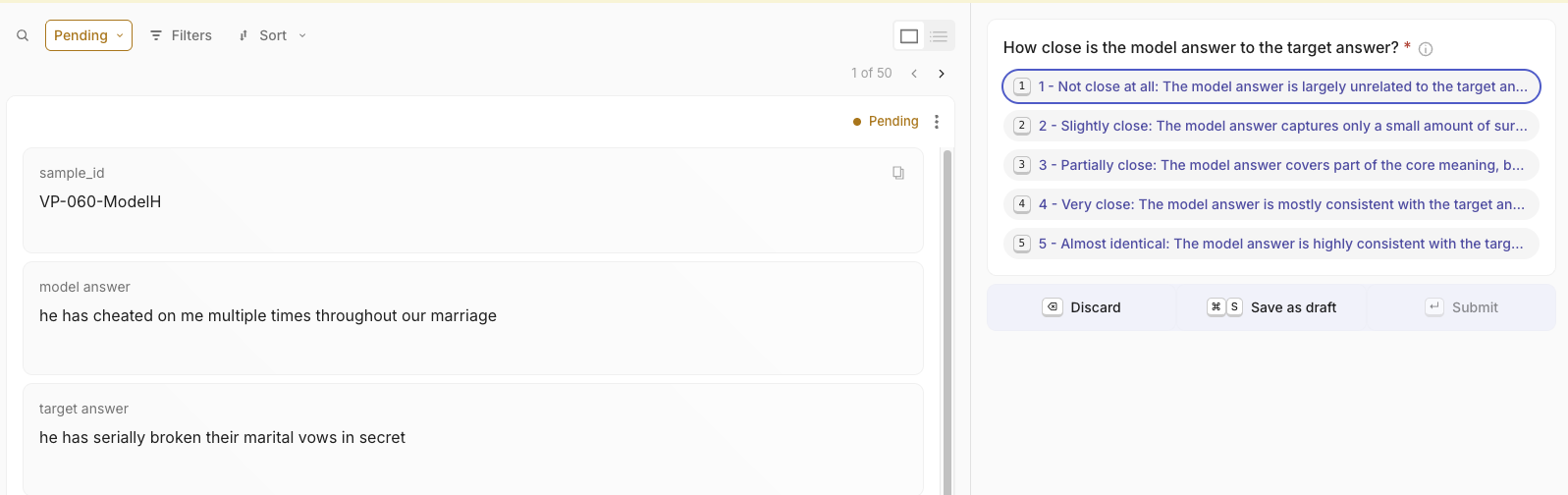}
    \caption{Screenshot of the human evaluation interface deployed with Argilla on Hugging Face.}
    \label{fig:human-evaluation-interface}
\end{figure*}

To examine whether the LLM-as-judge scores are consistent with human judgments, we conducted a human evaluation on two tasks: Emotion Recognition (ER) and Viewpoint Prediction (VP). For each task, we randomly sampled 100 prediction instances from the outputs of all evaluated models. The 100 samples from each task were further divided into two subsets of 50 samples, resulting in four annotation sets in total. We recruited four human evaluators online, all of whom were current graduate students at the master's level or above. The evaluation interface was implemented using Argilla\footnote{https://argilla.io} and deployed on Hugging Face. Each annotation set was independently distributed to one evaluator in a questionnaire-style format, where the evaluator was asked to compare the model-generated response with the corresponding dataset annotation and assign a similarity score from 1 to 5, with a higher score indicating stronger semantic consistency between the two answers. A screenshot of the annotation interface is shown in Figure~\ref{fig:human-evaluation-interface}. We then compared the human-assigned scores with the scores produced by the LLM-as-judge. As shown in Table~\ref{tab:human-evaluation}, the LLM-as-judge scores exactly matched the human scores in 159 out of 200 cases, corresponding to an exact agreement rate of 79.5\%. In addition, 35 cases differed by only one point, and 6 cases differed by two points, while no sample showed a discrepancy larger than two points. The overall mean absolute error (MAE) was 0.235, indicating that the LLM-as-judge scores were highly close to human ratings. The results are also consistent across the two tasks: for VP, the LLM-as-judge achieved 81 exact matches out of 100 samples with an average MAE of 0.20; for ER, it achieved 78 exact matches out of 100 samples with an average MAE of 0.27. Although ER exhibits slightly larger deviations than VP, all discrepancies remain within two score points. These results suggest that the LLM-as-judge evaluation closely aligns with human judgment and can serve as a reliable automatic scoring method for assessing the similarity between model predictions and annotated answers.

\section{Annotation Procedure}
\label{app:annotation_procedure}

This section describes the annotation procedure used to construct the relation-aware ESC benchmark.
Our annotation is a multimodal row-level process: subtitle text serves as the linguistic and temporal anchor, while audio and video provide additional evidence for emotion, interaction behavior, relational stance, and therapist intervention.
The overall procedure consists of four stages: preparing review assets, performing row-level multimodal annotation, conducting human review and revision, and exporting the final structured annotation files.

\subsection{Review Asset Preparation}
\label{app:review_asset_preparation}

We first manually split the original therapy videos into independent scene-level segments.
This step is performed by the authors mannually before row-level annotation.
A new scene segment is created when the main interactional focus changes, especially when the target participants or the relational unit under discussion changes.
This manual scene segmentation ensures that each segment contains a relatively coherent relational context and can be reviewed as a self-contained interaction episode.

After obtaining the scene-level segments, we prepare each segment as a reviewable multimodal asset.
Each segment contains the corresponding video clip and its aligned subtitle file.
The subtitle timestamps are used as the timing ground truth for spoken content.
Based on these timestamps, we further use the annotation model Gemini-3.1-Pro to create row-level video clips and corresponding subtitle snippets, so that each interaction row can be inspected independently while still being linked to the original scene context.

For each segment, we also create a structured manifest file that records the temporal span, subtitle text, speaker information, and available annotations for each row.
This manifest serves as the intermediate representation between raw multimodal data and the final benchmark annotations.
It allows annotators to review each row together with its surrounding context, instead of treating utterances as isolated text-only instances.

\begin{figure*}[t]
\centering
\begin{minipage}{\linewidth}
\begin{tcolorbox}[
  title={Example of annotation},
  colback=gray!5,
  colframe=gray!50,
  boxrule=0.5pt,
  arc=2pt
]
\scriptsize
\begin{verbatim}
{
  "sample_id": "couple_001",
  "rows": [
    ...
    {
      "row_index": 20,
      "start_time": 133.6,
      "end_time": 139.5,
      "primary_speaker": "Therapist",
      "target": ["Couple"],
    "dialogue_cleaned": "When you reach for him more urgently, he gets quieter, and then you feel even more alone and push harder.",
      "tone_of_voice": "slow, containing, non-blaming",
      "interaction_behavior": [
        {
          "initiator": "Therapist",
          "target": ["Couple"],
          "description": "alternates gaze between partners while mapping the sequence"
        }
      ],
      "support_strategy": [
        {
          "provider": "Therapist",
          "target": ["Couple"],
          "strategy_type": "track",
          "strategy_content": "maps the pursue-withdraw sequence without changing its meaning"
        }
      ],
      "relation_cycle_state": "pursue_withdraw",
        "relation_cycle_reason": "Mia presses for contact, while Josh lowers engagement, gives minimal answers, and withdraws.",
      "relation_cycle_evidence_rows": [18, 19, 20]
    },
    ...
  ]
}
\end{verbatim}
\end{tcolorbox}
\end{minipage}
\caption{Example of the row-level annotation format. The omitted rows are denoted by ellipses. Client rows include individual and relational understanding annotations, while therapist rows additionally include support-strategy and relation-pattern annotations.}
\label{fig:annotation_example}
\end{figure*}

\subsection{Annotation Unit}
\label{app:annotation_unit}

The basic annotation unit is a row-level interaction unit.
Each row is anchored to a specific time interval and represents a coherent interactional segment.
A new row is created when there is a change in speaker, addressee, interactional function, emotional meaning, or salient nonverbal behavior.
Therefore, the row segmentation is interaction-centered rather than purely subtitle-centered.

This means that a single subtitle span may be split into multiple rows if it contains multiple interactional units.
For example, if a speaker first responds to the therapist and then turns to address their partner, the segment may be divided into separate rows with different targets.
Likewise, if the same speaker shifts from explanation to accusation, or from defensive posture to visible softening, the segment may be split to preserve the change in relational function.
The purpose of this segmentation is not to produce a word-level transcript, but to obtain structured units that support relation-aware analysis.

\subsection{Annotated Dimensions}
\label{app:annotated_dimensions}

Each row is annotated along several dimensions.
First, timing and entity information records the start time, end time, primary speaker, and target participants.
Second, verbal content records the cleaned dialogue and utterance type.
Third, individual cues describe the speaker's tone of voice, facial expressions, body posture, self-directed behavior, and inferred internal emotion.
These fields provide evidence for understanding the participant's internal emotional state.

Beyond individual-level information, we annotate two relation-aware dimensions that support the benchmark tasks. 
First, relational stance records how participants orient toward one another, including interaction behavior and directed viewpoints or attitudes. 
Second, relation pattern summarizes the dominant relational cycle of the segment, together with a brief reason and supporting evidence rows. 
For therapist turns, we additionally annotate the therapist's primary support strategy and intervention intention. 
Figure~\ref{fig:annotation_example} illustrates the resulting row-level format: client rows mainly contain speaker grounding, verbal content, multimodal cues, internal emotion, and directed viewpoints, while therapist rows additionally include support-strategy annotations and relation-pattern evidence.

\subsection{Human Review and Revision}
\label{app:human_review}

After LLM-assisted pre-annotation, we build and deploy an online verification system for expert human review.
The system is designed to support row-level multimodal checking: annotators can browse prepared clips, inspect each segmented row, compare the draft annotation with the original video and subtitle evidence, revise incorrect fields, and submit the reviewed version back to the system.

\paragraph{Annotator recruitment.}
We recruit three PhD-level annotators to conduct the verification.
Before formal review, annotators are introduced to the label taxonomy, label-boundary rules, and the review interface.
They are instructed to treat the subtitle as the ground truth for spoken content and timing, while using the video and audio evidence to verify nonverbal cues, emotional states, interaction behavior, relation patterns, and therapist strategies.
The final labels used in the benchmark are obtained after this human verification stage, rather than directly from the LLM-generated draft.

\paragraph{System deployment.}
To support human verification, we implement the review interface as a web-based annotation system and deploy it on a Linux server within the local area network.
The prepared review assets, including video clips, aligned subtitles, row-level clips, manifest files, and draft annotations, are stored on the server and served through the internal website.
Annotators access the system through a browser on the same local network, which allows them to review the data without manually transferring large video files.
This deployment also ensures that all edits are written back to a centralized location, making it possible to track progress, collect revised annotations, and manage reviewed versions consistently.

\begin{figure*}[t]
  \centering
  \includegraphics[width=\linewidth]{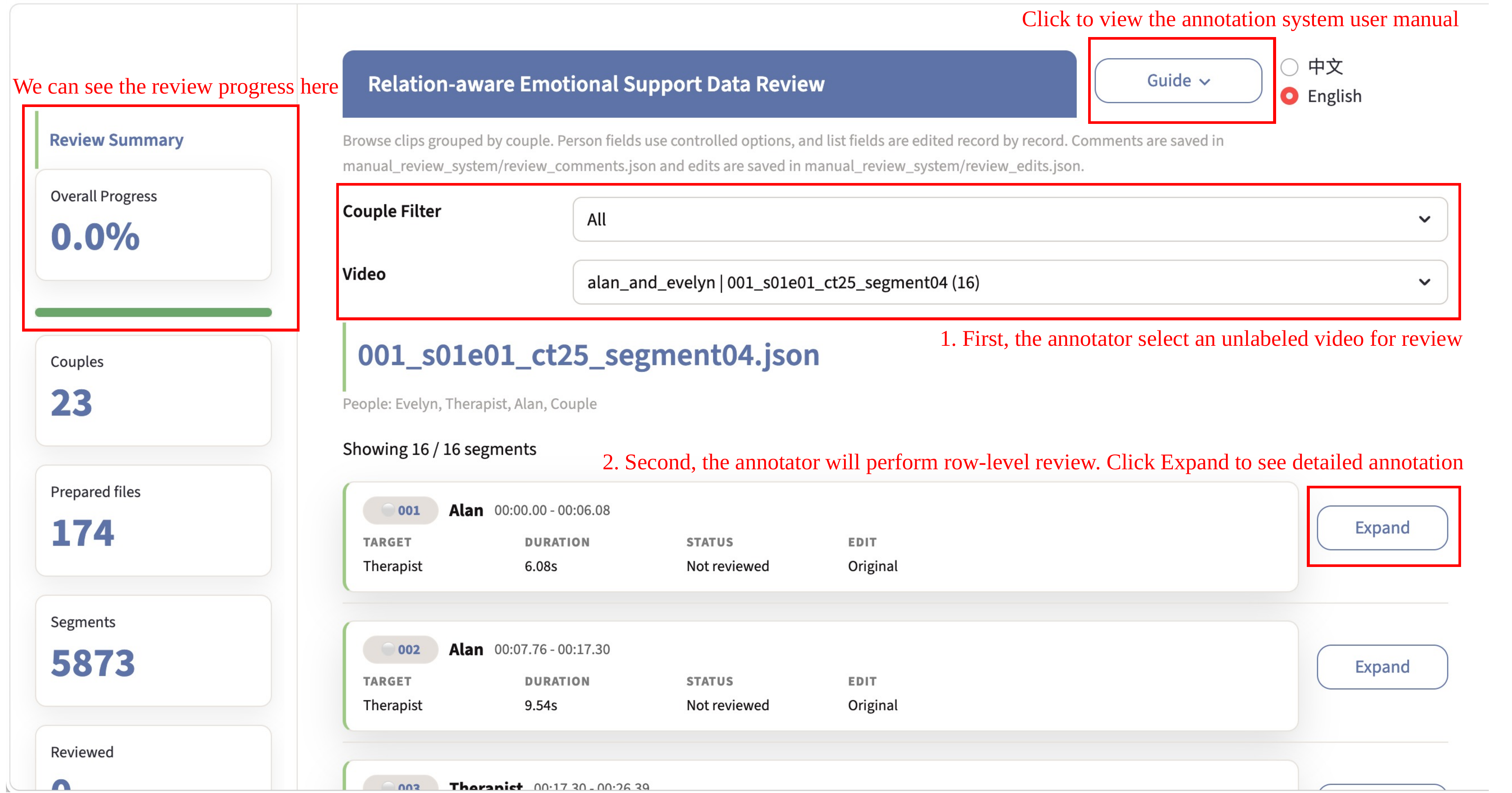}
  \caption{
  Overview of the web-based annotation review dashboard.
  The left panel summarizes review progress, including prepared files, total segments, reviewed rows, remaining rows, and commented segments.
  Annotators select an unreviewed video segment from the dropdown menus and expand each row to inspect its detailed annotation.
  }
  \label{fig:annotation_process_1}
\end{figure*}

\paragraph{Review dashboard.}
As shown in Figure~\ref{fig:annotation_process_1}, the review system provides a dashboard for browsing and managing annotation progress.
Annotators can view the overall review status, including the number of prepared files, total segments, reviewed rows, remaining rows, and commented segments.
They first select an unreviewed video segment from the video list and then perform row-level review.
Each row is displayed with its speaker, target, duration, review status, and edit status.
By clicking the expand button, annotators open the detailed annotation view for that row.

\begin{figure*}[t]
  \centering
  \includegraphics[width=\linewidth]{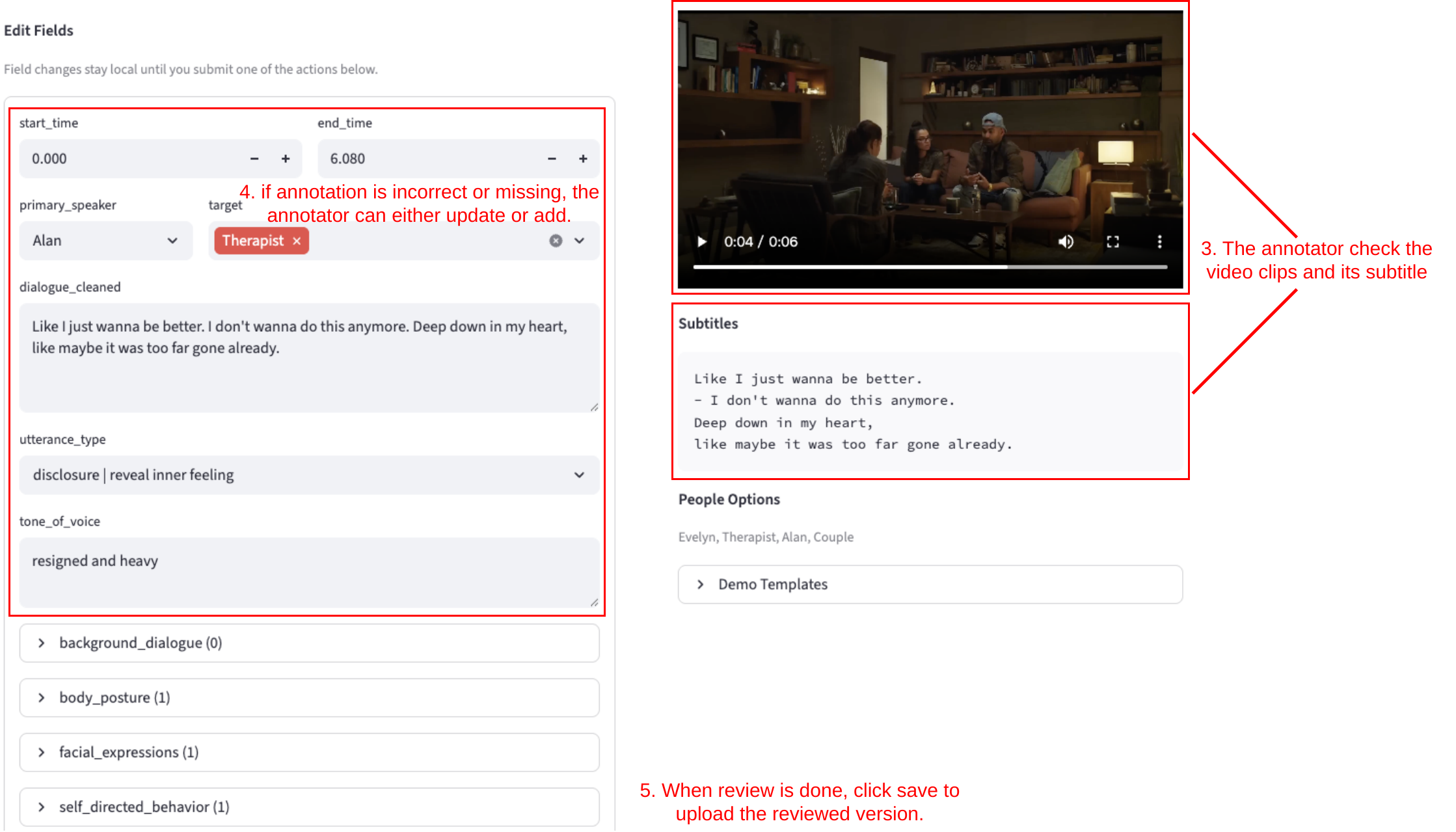}
  \caption{
  Row-level verification interface.
  Annotators review the original video clip and aligned subtitles, compare them with the structured annotation fields, and revise or add missing information when necessary.
  After checking the row-level evidence, annotators save the reviewed version, which is uploaded to the centralized server.
  }
  \label{fig:annotation_process_2}
\end{figure*}

\paragraph{Row-level verification.}
As shown in Figure~\ref{fig:annotation_process_2}, the detailed row-level interface presents the video clip, aligned subtitles, participant options, and editable annotation fields side by side.
Annotators first watch the video clip and read the subtitle to verify the spoken content, timing, speaker, and target.
They then inspect the structured fields, including utterance type, tone of voice, body posture, facial expressions, self-directed behavior, interaction behavior, internal emotion, viewpoints, relation pattern, and therapist strategy when applicable.
If a field is incorrect, incomplete, or missing, annotators can directly revise the existing value or add a new entry in the interface.
For example, they may correct a mistaken target, revise an emotion description, add missing posture evidence, change a relation-pattern label, or replace a support-strategy label that reflects surface wording rather than the therapist's primary intervention function.

\paragraph{Submission and filtering.}
Edits remain local in the interface until the annotator explicitly saves the reviewed row or segment.
After completing the review, the annotator clicks the save button to upload the reviewed version to the server.
Segments are discarded when reliable verification is not possible, including cases with severe subtitle errors, unresolved speaker mismatches, missing or ambiguous video evidence, incomplete context, or insufficient relational information.
This process ensures that the released benchmark annotations are grounded in the original multimodal evidence and have been manually checked before being converted into task-specific evaluation instances.

In future work, we plan to open-source the annotation interface and invite broader community participation under appropriate ethical, copyright, and data-use constraints.

\clearpage
\onecolumn

\section{Details of Prompts}
\subsection{Prompts for LLM-based Pre-annotation}
\label{app:preannotation_prompts}

This section provides the full prompts used for LLM-based pre-annotation.
We use two scenario-specific prompts: one for couple conversations and one for family conversations.
Both prompts instruct the multimodal LLM Gemini-3.1-Pro to perform row-level annotation using subtitle text as the verbal ground truth and audio-visual evidence as additional support for emotion, posture, interaction behavior, relational stance, and therapist intervention.
The prompts are included verbatim below.

\subsubsection{Couple Pre-annotation Prompt}
\label{app:couple_preannotation_prompt}

\begin{tcolorbox}[
  colback=gray!5,
  colframe=gray!50,
  boxrule=0.5pt,
  arc=2pt,
  breakable,
  left=4pt,
  right=4pt,
  top=4pt,
  bottom=4pt
]
\begin{Verbatim}
### Role: Senior Clinical Interaction Analyst & Multimodal Transcript Annotator

Objective:
Perform high-fidelity multimodal annotation of the provided therapy video using subtitle text as the verbal ground truth.
Return JSON only.

The output must follow this exact top-level schema:
{
  "rows": [
    {
      "start_time": 0.0,
      "end_time": 0.0,
      "primary_speaker": "",
      "target": [],
      "dialogue_cleaned": "",
      "utterance_type": "",
      "background_dialogue": [],
      "tone_of_voice": "",
      "body_posture": [],
      "facial_expressions": [],
      "self_directed_behavior": [],
      "interaction_behavior": [],
      "internal_emotion": [],
      "viewpoints_attitudes": [],
      "support_strategy": []
    }
  ]
}

Hard output rules:
1. Return valid JSON only.
2. Do not wrap JSON in Markdown or code fences.
3. Do not add keys, rename keys, or omit keys.
4. Use [] for empty array fields.
5. Do not use null, None, N/A, placeholder labels, or free-form notes outside the schema.
6. Do not create any record with an empty `source`, empty `target`, or empty target array.

---

## Entity Naming Rules

This annotation uses a CLOSED entity set for all person/object fields.

Allowed entity labels are ONLY:
1. the visible/interacting interview participant names in this clip
2. `Therapist`
3. `Couple`

These rules apply to ALL person-bearing fields:
- `primary_speaker`
- `target`
- `background_dialogue[].speaker`
- `background_dialogue[].target`
- `body_posture[].person`
- `facial_expressions[].person`
- `self_directed_behavior[].person`
- `interaction_behavior[].initiator`
- `interaction_behavior[].target`
- `internal_emotion[].person`
- `viewpoints_attitudes[].source`
- `viewpoints_attitudes[].target`
- `support_strategy[].provider`
- `support_strategy[].target`

Forbidden labels:
- any self-referential placeholder label
- any group alias other than `Couple`
- any therapist personal name
- any off-screen or historical person
- any invented placeholder such as artificial listener labels

Normalization rules:
- Always use `Therapist` for the practitioner, never a therapist personal name.
- Always use `Couple` for speech or behavior genuinely directed to all interview participants at once.
- Never create a new person label for an off-screen person who is merely mentioned in the dialogue.
- If someone reflects on themselves, use their actual on-screen name only in non-target person fields such as `internal_emotion[].person`, `self_directed_behavior[].person`, `body_posture[].person`, or `facial_expressions[].person`.
- Do not use a person as their own `target` anywhere in the schema.

Important:
- A clip may contain two participants or three participants. Do not assume there are exactly two.
- If three interview participants are present, use their actual names individually when the target/source is specific.
- Use `Couple` only when the utterance or behavior is truly directed to the group as a whole.
- Special rule for `support_strategy`:
  - `support_strategy` is therapist-only
  - use only the closed strategy set defined below
  - follow the detailed `support_strategy` constraints below exactly

---

## Segmentation Rules

Create a new row whenever any of the following changes:
1. speaker
2. target
3. interactional unit
4. meaningful emotional or relational shift
5. meaningful nonverbal interaction shift

Hard segmentation rule:
If the same speaker sequentially addresses different targets within one subtitle span, split into multiple rows and split the dialogue accordingly.

Do not merge speech across different targets.

---

## Subtitle Integrity Rules

1. `dialogue_cleaned` must stay faithful to the subtitle text.
2. Do not invent dialogue.
3. Do not paraphrase spoken words.
4. You may remove subtitle formatting prefixes such as `Josh:` or `[Therapist]` when they are labels rather than spoken content.
5. If a name is actually spoken in the utterance, keep it in `dialogue_cleaned`.
6. `background_dialogue[].content` must also remain subtitle-faithful when available.

## Subtitle Coverage And Timing Consistency Rules

1. Treat the subtitle file as the timing ground truth for spoken content.
2. Every spoken subtitle line with lexical content must be accounted for in either:
   - `dialogue_cleaned`, or
   - `background_dialogue[].content`
3. Do not silently drop subtitle lines just because they are short, awkward, overlapping, or emotionally minor.
4. If a subtitle span contains two distinct spoken lines that belong in different rows, split them across rows, but still preserve both lines.
5. Times must be computed in seconds from the subtitle timestamps exactly. Never drop the minute component.
6. Example: `00:01:40,000` is `100.0` seconds, not `40.0`.
7. Every row's `start_time` and `end_time` must stay close to the subtitle span that supports that row.
8. The final annotated `end_time` must not exceed the source clip duration.
9. The annotation must continue far enough to cover the final spoken subtitle line in the clip.
10. The source subtitle/video timing intervals are read-only evidence. Do not rewrite, relocate, stretch, or compress a spoken line into a different time region just to make the annotation look cleaner.
11. Do not invent synthetic timing windows that are not supported by the source subtitles. If a short line such as `Yeah.` or `Cool.` lives inside one subtitle cue, keep it inside that real source interval rather than moving it into a new adjacent gap.
12. Do not move a line forward or backward into the next speaker's interval, the previous speaker's interval, or a silence gap. If the timing evidence is one cue or one continuous cue window, keep the row anchored to that same source region.

---

## Target Resolution Rules

Determine `target` by dialogue logic, not by name mention alone.

Use this priority order:
1. direct response target
2. direct address target
3. ongoing conversational exchange target
4. visible gaze/body orientation target
5. group target only when truly collective

Examples:
- If a participant talks to the therapist about their partner, target is `["Therapist"]`, not the partner.
- If the therapist addresses all interview participants together, target is `["Couple"]`.
- If a participant blurts, sighs, laughs, tears up, or reacts without clear direct address, still assign a non-empty top-level `target` based on the ongoing exchange context, usually the current conversational recipient, most salient addressee, or current speaker they are reacting to.
- In those self-directed reaction cases, still represent the speaker-centered reaction through fields such as `self_directed_behavior`, `facial_expressions`, `body_posture`, and `internal_emotion`; do not use a self-target.
- If one participant speaks directly to two named participants at the same time, include both names in the target array.

---

## Field Definitions

Coverage policy for sparse fields:
- The following fields are often under-filled by weaker models and therefore require an explicit scan on every row:
  - `facial_expressions`
  - `self_directed_behavior`
  - `interaction_behavior`
  - `internal_emotion`
  - `viewpoints_attitudes`
  - `support_strategy`
- For EVERY row, actively check whether each of these six fields has at least one justified annotation.
- Do not leave one of these fields empty merely because the evidence is subtle. If there is reasonable multimodal evidence, add a short conservative record.
- Leave a field empty only when there is truly no observable or inferable evidence for that field in that row.
- In therapy footage, emotionally or relationally meaningful rows should often have at least one item in `internal_emotion`, `viewpoints_attitudes`, or `support_strategy`.
- Nonverbal reaction rows with little or no speech should still be annotated through the nonverbal fields when possible.

### 1. start_time
- number in seconds
- must be anchored to the real subtitle/video interval that supports the row
- do not rewrite the line into an earlier or later region than the source evidence

### 2. end_time
- number in seconds
- must be greater than `start_time`
- must stay within the real supporting subtitle/video interval or continuous cue window for that row
- do not extend the row into silence, into the next unsupported interval, or past the source clip duration

### 3. primary_speaker
- string
- must be one allowed entity label

### 4. target
- array of strings
- every item must be one allowed entity label
- even one target must still be an array
- `target` must not be empty
- `target` must not include `primary_speaker`

### 5. dialogue_cleaned
- string
- main foreground utterance only

### 6. utterance_type
- string
- this field is ONLY for non-`Therapist` rows
- if `primary_speaker` is `Therapist`, set `utterance_type` to the empty string `""`
- if `primary_speaker` is not `Therapist`, choose exactly one from:
  - `accusation`
  - `defense`
  - `disclosure`
  - `request`
  - `recollection`
  - `reflection`
  - `repair`
  - `withdrawal`
  - `validation`
  - `other`
- choose the MAIN utterance function, not the emotion
- definitions:
  - `accusation`: blame, criticize, or assign fault to another person
  - `defense`: justify, rebut, deny, or protect oneself against criticism
  - `disclosure`: reveal inner feeling, vulnerability, fear, hurt, or need
  - `request`: ask for a response, change, reassurance, or concrete need
  - `recollection`: narrate or recall a past event, example, or background
  - `reflection`: articulate insight or self-understanding about one's own pattern or reaction
  - `repair`: soften, apologize, clarify goodwill, or reconnect
  - `withdrawal`: shut down, disengage, deflect, or refuse entry
  - `validation`: affirm or receive another person's feeling or experience
  - `other`: use only if none of the above fits

### 7. background_dialogue
- array of objects with exact schema:
{
  "speaker": "",
  "target": [],
  "content": ""
}

Rules:
- `speaker` must be one allowed entity label
- every `target` item must be one allowed entity label
- `target` must not be empty
- `speaker` must not appear in `target`
- use only for overlapping or secondary speech
- If overlapping speech is only a laugh, sigh, gasp, or short self-directed vocal reaction and no real outward target can be identified, do not force a `background_dialogue` record; use self-focused fields instead.
- If you cannot assign a real non-empty outward `target`, do not create a `background_dialogue` record for that reaction.

### 8. tone_of_voice
- string
- short descriptive phrase grounded in audible delivery

### 9. body_posture
- array of objects:
{
  "person": "",
  "description": ""
}

Rules:
- `person` must be one allowed entity label
- posture/orientation only, not gestures

### 10. facial_expressions
- array of objects:
{
  "person": "",
  "description": ""
}

Rules:
- `person` must be one allowed entity label
- visible facial evidence only
- Actively annotate this field whenever a face shows a readable expression such as smiling, smirking, brow furrowing, grimacing, tearing up, tightening, widening eyes, blanking out, or looking pained/exasperated.
- If the expression changes the interpersonal meaning of the moment, include it.
- Do not leave empty if a visible facial reaction is clearly doing interactional work.

### 11. self_directed_behavior
- array of objects:
{
  "person": "",
  "description": ""
}

Rules:
- `person` must be one allowed entity label
- use for self-directed or not-clearly-targeted actions
- gestures toward another person belong in `interaction_behavior`
- Actively use this field for sighing, laughing, crying, rubbing face, touching forehead, looking away, shaking head to oneself, fidgeting, collapsing into posture, pausing to regulate, or other self-managed reactions.
- If the behavior is primarily about the person managing themselves rather than acting on someone else, put it here.
- For laughter, sighing, crying, head-shaking to oneself, or similar self-directed reactions, use this field with `person = primary_speaker` rather than making that person their own `target`.

### 12. interaction_behavior
- array of objects:
{
  "initiator": "",
  "target": [],
  "description": ""
}

Rules:
- `initiator` must be one allowed entity label
- every `target` item must be one allowed entity label
- `target` must not be empty
- `initiator` must not appear in `target`
- must describe observable interpersonal physical/nonverbal behavior only
- do not put abstract intentions or interpretations here
- Actively annotate this field for eye contact, gaze shifts, turning toward someone, interruptive leaning, pointing, reaching, withdrawing from someone, orienting body toward someone, laughing directly at someone, or other visible relational moves between people.
- If a nonverbal act is clearly directed at another on-screen person or at the whole group, prefer annotating it here rather than leaving it empty.
- If the behavior is not clearly directed outward to another person or the group, do not force an `interaction_behavior` record; use `self_directed_behavior`, `facial_expressions`, `body_posture`, or `internal_emotion` instead.

### 13. internal_emotion
- array of objects:
{
  "person": "",
  "emotion_description": "",
  "intensity": 0
}

Rules:
- `person` must be one allowed entity label
- `intensity` must be a number from 0 to 10
- use conservative, evidence-based affect inference only
- This field should be filled whenever emotion is reasonably inferable from tone, wording, pacing, posture, facial expression, or behavior.
- Prefer specific phrases such as `feels defensive and cornered`, `feels warmly encouraged`, `feels resigned and tired`, `feels amused but dismissive` instead of vague labels like `sad` or `mad`.
- If the row is emotionally charged, do not leave this empty unless the evidence is genuinely unreadable.

### 14. viewpoints_attitudes
- array of objects:
{
  "source": "",
  "target": "",
  "viewpoint": "",
  "attitude": ""
}

Rules:
- `source` must be one allowed entity label
- `target` must be one allowed entity label and must be different from `source`
- do not use off-screen or historical people as `target`
- `viewpoint` and `attitude` must be short descriptive phrases
- Use this field only for attitudes toward another on-screen person, not for self-evaluation or self-reflection.
- Actively annotate this field whenever someone expresses blame, admiration, criticism, distrust, defensiveness, disappointment, dismissal, longing, respect, frustration, or a stable stance toward another on-screen person.
- If a speaker is making a claim about who another person is, what they do, how they fail, what they intend, or how they are experienced, that usually belongs here.
- Therapy dialogue frequently contains explicit attitudes and implicit relational positions; capture them conservatively rather than leaving this field empty by default.
- If no clear outward target exists, do not create a `viewpoints_attitudes` record; put self-focused material into `internal_emotion` or other non-target fields instead.

### 15. support_strategy
- array of objects:
{
  "provider": "",
  "target": [],
  "strategy_type": "",
  "strategy_content": ""
}

Rules:
- `provider` must always be `Therapist`
- every `target` item must be a real interview participant name or `Couple`
- `target` must never include `Therapist`
- `provider` must not appear in `target`
- if `Couple` is used, it must be the only target item
- annotate only genuine supportive moves
- We encourage annotating `support_strategy` whenever the CURRENT therapist turn clearly performs a real supportive intervention. Do not leave it empty just because you want to be conservative.
- Do not create a `support_strategy` record unless the therapist's support has a clear non-empty target person or `["Couple"]`.
- Hard rule: minimal acknowledgments, backchannels, neutral continuers, and ordinary clarification/content/cause questions are empty by default and should NOT be annotated as `support_strategy` unless the CURRENT turn itself clearly performs a closed-set intervention.
- Do not infer a strategy from surrounding context when the CURRENT turn itself is only `yeah`, `right`, `okay`, `i see`, `uh-huh`, `oh`, `wow`, or a thin clarification such as `what happened?` or `how so?`.
- Simple speaker handoff prompts such as `and you?`, `what about you?`, or `how about you?` are empty by default unless the CURRENT turn is clearly repairing a live asymmetry in attention or alliance.
- Cause-reconstruction or fact-finding questions such as `what made X suspicious`, `what happened next`, `what kind of company`, `when did that happen`, or `who was there` are empty by default unless the CURRENT turn itself explicitly names the live interaction pattern or shifts the meaning of the conflict.
- Brief emotion-check follow-ups are not automatically strategy labels. Use `evoke` only when the CURRENT turn clearly distills a deeper primary emotion or vulnerability, not when it merely echoes or lightly probes already-stated affect.
- These labels overlap conceptually. Choose the MAIN FUNCTION of the CURRENT turn, not every plausible function.
- Prefer the most specific justified label. `track` is the broadest active label and should win only when no more specific label clearly fits.
- closed strategy types:
  - `counterbalance`
  - `safeguard`
  - `goal_align`
  - `track`
  - `reframe`
  - `evoke`
  - `enact`
  - `join`
  - `detach`
  - `repair`
- Use these types in a COUPLE-THERAPY / CONJOINT-THERAPY sense rather than as generic one-person comfort labels.
- No single label is preferred. Do not default to `track` or any other one category when the turn is ambiguous.
- These labels are anchored in couple/family therapy intervention literature:
  - `counterbalance`: counterbalancing moves / balancing alliances; brings a sidelined partner back into the interaction after asymmetry
  - `safeguard`: safety within the therapeutic system; protects a vulnerable person from being overrun, shamed, or overwhelmed
  - `goal_align`: shared sense of purpose; helps the couple orient to a common task, focus, or therapeutic goal
  - `track`: tracking interactions; explicitly follows the live negative cycle or interaction pattern between partners
  - `reframe`: reframing; shifts blame into a relational frame, interaction cycle, or shared dilemma
  - `evoke`: evocative responding; invites deeper vulnerability, fear, hurt, longing, or need
  - `enact`: enactment; helps one person say something directly to another person in the room
  - `join`: empathic joining; helps partners connect through softer, more vulnerable emotional disclosure
  - `detach`: unified detachment; helps the couple step back and look together at the problem or pattern rather than fighting as opponents
  - `repair`: attachment injury repair / rupture repair; softens rupture, clarifies intent, apologizes, or rebuilds connection
- Important distinctions:
  - `safeguard` is about SAFETY. Use it when the main function is protecting a vulnerable person or making the room safe enough for them to stay engaged.
  - `counterbalance` is about BALANCE. Use it when the main function is restoring symmetry, bringing the less-heard partner back in, or correcting momentary siding/asymmetry.
  - `track` is about naming the live interaction sequence, cycle, contradiction, or recurring pattern.
  - `reframe` goes beyond tracking and changes the meaning of the problem into a relational pattern, shared dilemma, or different interpretive frame.
  - `evoke` draws out deeper primary emotion, fear, hurt, shame, longing, or need; simple surface affect labeling is usually not enough.
- `join` uses softer vulnerable contact to increase emotional receptivity or closeness between participants; warmth or praise alone is not enough.
- `enact` is specifically about getting one person to say or hear something directly to/from another person in-session.
- `detach` helps both partners take a shared observer stance toward the problem; it is more than ordinary pattern summary.
- `goal_align` explicitly reorients the participants toward what they are trying to do together, the shared task, or the next joint focus; it is not generic summarizing.
- `repair` is about softening an existing rupture, receiving an apology, clarifying goodwill, or restoring connection after strain; generic soothing is not enough.
- Do NOT fall back to generic labels like validation, perspective-taking, or emotional soothing. Always choose the closest label from the closed set above.
- `track`, `reframe`, and `evoke` are not fallback labels. If the current turn is weak, generic, or purely facilitative, leave `support_strategy` empty.
- If two or more labels look possible, resolve the boundary by choosing the most specific function carried by the CURRENT turn, not the broadest one.
- The goal is not to minimize labels. The goal is to label genuine therapist strategy moves while keeping weak or merely clarifying turns empty.
- Only annotate this field when the CURRENT therapist turn itself clearly restores alliance balance, protects vulnerability, builds shared purpose, tracks the cycle, reframes conflict, deepens vulnerability, coaches direct partner-to-partner contact, fosters empathic joining, creates unified detachment, or repairs relational injury.
- Therapist questions are often not neutral, but ordinary clarification is still empty by default. Only annotate a question as `support_strategy` when the CURRENT question clearly tracks, reframes, evokes, enacts, safeguards, counterbalances, aligns goals, or repairs disconnection.
- Boundary reminders:
  - `safeguard` wins over `counterbalance` when protection/safety is the main function
  - `counterbalance` wins over `track` when the main move is restoring participation or alliance balance
  - `repair` wins over `join` when the main move is repairing rupture, receiving apology, or re-establishing connection after strain
  - `enact` wins over `evoke` when the therapist is explicitly moving one person toward direct in-room communication
  - `goal_align` wins over `track` when the main move is reorienting the couple toward a shared task or purpose
  - `detach` wins over `track` when the main move is helping both people step back together and observe the problem as a shared pattern
  - `reframe` wins over `track` when the turn changes the meaning of the problem rather than merely naming the sequence
  - `track` names the live sequence or cycle; ordinary clarification is not `track`
  - asking for the reason, cause, trigger, chronology, or factual setup of a story is not enough for `track`
  - if a turn could be read as either a broad `track` move or a more specific move such as `counterbalance`, `evoke`, `reframe`, `enact`, or `repair`, prefer the more specific label when the CURRENT turn supports it
  - `reframe` changes the meaning of the problem; blunt judgment alone is not enough
  - `evoke` draws out deeper primary emotion; generic prompting, surface emotion echoing, or brief follow-up like `for being upset?` is not enough
  - `counterbalance` restores asymmetry in airtime or alliance; simple turn-taking or `and you?` is not automatically `counterbalance`
  - `safeguard` protects a vulnerable person from being overrun; generic agreement is not `safeguard`
  - `join` deepens softer connection; simple praise or warmth alone is not always `join`
  - `detach` requires a shared observer stance; therapist-only pattern summary is often still `track` or `reframe`
  - `goal_align` requires explicit shared-task orientation; generic process summary is not `goal_align`
  - `repair` softens rupture or reconnects after strain; do not silently fold all such moments into `evoke`, `join`, or `reframe`
- Example defaults:
  - `You're hurt?` may be `evoke` if it clearly names deeper primary emotion
  - `For being upset?` is usually empty
  - `So what made Evelyn suspicious of you?` is usually empty
  - `And you?` is usually empty unless it clearly repairs asymmetry
- Even when a partner is being warm, helpful, or reparative, do NOT annotate it in `support_strategy` unless the support is clearly being delivered by `Therapist`. This field is only for therapist-originated strategies.

---

## Additional Strict Constraints

1. Do not output any illegal entity labels anywhere in the JSON.
2. Do not use therapist personal names in any structured field; always use `Therapist`.
3. Do not use self-referential placeholders anywhere in the JSON.
4. Do not use any group alias other than `Couple`.
5. Do not use off-screen family members, ex-partners, or other historical figures as structured entities.
6. When off-screen people are discussed, keep them only inside free-text descriptive fields such as:
- `dialogue_cleaned`
- `tone_of_voice`
- `emotion_description`
- `viewpoint`
- `attitude`
- `strategy_content`
7. Use descriptive short phrases rather than one-word tags whenever evidence supports it.
8. Keep all annotations evidence-based and multimodally grounded.
9. Before finalizing a row, explicitly ask whether each of the six sparse fields can be filled conservatively rather than defaulting to `[]`.
10. `utterance_type` must be `""` for `Therapist` rows and one allowed label for non-`Therapist` rows.
11. Every `target` array in the schema must be non-empty.
12. In every field that has an actor/source/provider and a target, those must refer to different people.
13. For optional target-based sub-records such as `background_dialogue`, `interaction_behavior`, `viewpoints_attitudes`, or `support_strategy`, if no real outward target can be identified, use self-focused fields such as `self_directed_behavior`, `facial_expressions`, `body_posture`, or `internal_emotion` instead of forcing that sub-record.
14. Do not leave meaningful subtitle content uncovered. If a spoken subtitle line exists, it must appear somewhere in the annotation unless it is purely non-lexical or unintelligible.
15. Do not output timing that implies minute-to-second collapse, such as converting material around `01:40` into the `40` second range.

---

## Final Self-Check Before Output

Before returning JSON, verify:
1. Every person/object field uses only allowed labels.
2. No self-referential placeholder, non-`Couple` group alias, or therapist personal name remains anywhere in structured entity fields.
3. No off-screen person is used as `primary_speaker`, `target`, `source`, `provider`, `initiator`, or `person`.
4. All array fields are arrays.
5. All rows contain all required keys.
6. For every row, you explicitly checked the six sparse fields and only left them empty when unsupported.
7. If a row contains clear emotion, stance, supportive guidance, or visible nonverbal behavior, at least the relevant sparse fields are populated.
8. Every `support_strategy[].provider` is exactly `Therapist`, and no `support_strategy[].target` contains `Therapist`.
9. Every `utterance_type` follows the therapist/non-therapist rule exactly.
10. No `target` array is empty.
11. `primary_speaker` is not in `target`; `background_dialogue[].speaker` is not in `target`; `interaction_behavior[].initiator` is not in `target`; `viewpoints_attitudes[].source != target`; and `support_strategy[].provider` is not in `target`.
12. The spoken subtitle coverage is materially complete: meaningful subtitle lines were not dropped from the annotation.
13. No row shows obvious minute/second arithmetic collapse; subtitle times like `01:40` were converted to about `100` seconds, not `40`.
14. The final annotated `end_time` does not exceed the source clip duration and still reaches the last spoken subtitle line.
15. No row rewrites the source timing interval: a line was not moved into an earlier/later region, stretched across an unsupported gap, or split into synthetic time windows that the subtitles do not support.

## Input
The following is the subtitle text:
\end{Verbatim}
\end{tcolorbox}

\clearpage

\subsubsection{Family Pre-annotation Prompt}
\label{app:family_preannotation_prompt}

\begin{tcolorbox}[
  colback=gray!5,
  colframe=gray!50,
  boxrule=0.5pt,
  arc=2pt,
  breakable,
  left=4pt,
  right=4pt,
  top=4pt,
  bottom=4pt
]
\VerbatimInput[
  fontsize=\scriptsize,
  breaklines=true,
  breakanywhere=true,
  tabsize=2
]{prompts/preannotation_family_prompt.txt}
\end{tcolorbox}

\subsection{Prompts for relation-aware ESC tasks}
\label{app:task_prompt}
This section presents the prompts used for zero-shot evaluation on the six relation-aware ESC tasks. The prompts are designed to keep the task definitions consistent across all evaluated models, while allowing different task types to use appropriate output formats. Specifically, ER and VP use generation-style outputs, ITP and RPP use classification-style outputs, and STP and SSP use ranking-style outputs. All models are evaluated with the same prompt templates and the same candidate label sets.

\subsubsection{Prompt for the ITP task}

\begin{tcolorbox}[
  colback=gray!5,
  colframe=gray!50,
  boxrule=0.5pt,
  arc=2pt,
  breakable,
  left=4pt,
  right=4pt,
  top=4pt,
  bottom=4pt
]
\begin{Verbatim}
You are evaluating therapist intervention timing in a couples-therapy or family-therapy interaction. For each candidate decision point, decide whether the therapist should take the next turn immediately after decision_after_row. Return yes if a therapist intervention is timely and clinically helpful now, and return no if the therapist should keep listening and let the clients continue. Use a concise clinical standard: prefer yes when the process has crystallized enough that a question, reflection, validation, containment, reframing, or process intervention would likely help now, and prefer no when a speaker is still unfolding a thought, vulnerable material is still emerging, or the exchange is still productively developing without therapist interruption. Judge only from the provided context and do not assume access to the unseen next turn.
\end{Verbatim}
\end{tcolorbox}

\subsubsection{Prompt for the RPP task}

\begin{tcolorbox}[
  colback=gray!5,
  colframe=gray!50,
  boxrule=0.5pt,
  arc=2pt,
  breakable,
  left=4pt,
  right=4pt,
  top=4pt,
  bottom=4pt
]
\begin{Verbatim}
You are predicting the dominant couple or family relation state at a target turn in a therapy interaction. The target is the dyadic or systemic interaction pattern organizing the interaction at that moment, not just one person's inner emotion. Predict only from the provided evidence and do not assume access to future rows. Focus on the dominant organization at the target turn being predicted; use labels such as repair_softening, constructive_alignment, cooperative_family_alliance, or mixed_transition only when their defining relational conditions are clearly present. Return exactly three distinct relation labels from the closed-set taxonomy, ranked from most to least likely, together with a short English reason grounded in the provided evidence.
\end{Verbatim}
\end{tcolorbox}

\subsubsection{Prompt for the SSP task}

\begin{tcolorbox}[
  colback=gray!5,
  colframe=gray!50,
  boxrule=0.5pt,
  arc=2pt,
  breakable,
  left=4pt,
  right=4pt,
  top=4pt,
  bottom=4pt
]
\begin{Verbatim}
You are predicting the therapist's next supportive strategy in a couples-therapy or family-therapy interaction. You only see prior context before the target therapist turn and must not assume access to the upcoming therapist utterance. The support target is given as known_support_target, and your task is only to rank the three most likely therapist strategies from the closed set. Predict the main function of the likely next therapist move, not every plausible function, and prefer the most specific justified label; broad labels such as track should win only when no more specific label clearly fits. Because the support target is given, use it to disambiguate between overlapping strategies, and return exactly three distinct strategy labels together with a short reason based only on the prior context and the known support target.
\end{Verbatim}
\end{tcolorbox}

\subsubsection{Prompt for the STP task}

\begin{tcolorbox}[
  colback=gray!5,
  colframe=gray!50,
  boxrule=0.5pt,
  arc=2pt,
  breakable,
  left=4pt,
  right=4pt,
  top=4pt,
  bottom=4pt
]
\begin{Verbatim}
You are predicting the therapist's next supportive target in a couples-therapy or family-therapy interaction. You only see prior context before the target therapist turn and must not assume access to the upcoming therapist utterance. Your task is to rank the three most likely support targets from the closed candidate target labels. Predict who the therapist's next supportive move is mainly directed toward, use the group label (Couple or Family) when the likely move is jointly directed to the whole participant group or a shared pattern, and prefer a specific person when the likely move primarily invites, protects, challenges, reassures, or redirects one participant. Use relational flow, emotional activation, withdrawal or pursuit patterns, floor-taking, and the therapist's ongoing focus in the prior context.
\end{Verbatim}
\end{tcolorbox}

\subsubsection{Prompt for the ER task}

\begin{tcolorbox}[
  colback=gray!5,
  colframe=gray!50,
  boxrule=0.5pt,
  arc=2pt,
  breakable,
  left=4pt,
  right=4pt,
  top=4pt,
  bottom=4pt
]
\begin{Verbatim}
You are evaluating single-user understanding in a therapy session. Your task is: once the current speaker begins speaking, predict that speaker's internal emotion and intensity at that moment. Focus only on the row's primary_speaker; do not switch to the partner, therapist, or the dyad as the prediction target. Do not use future rows or outside context, because this task is intentionally about the current speaking turn itself. Return short, concrete emotion phrases such as "feels dismissed and frustrated" or "feels ashamed and cornered," and predict intensity as an integer from 0 to 10, with a short English reason.
\end{Verbatim}
\end{tcolorbox}

\subsubsection{Prompt for the VP task}

\begin{tcolorbox}[
  colback=gray!5,
  colframe=gray!50,
  boxrule=0.5pt,
  arc=2pt,
  breakable,
  left=4pt,
  right=4pt,
  top=4pt,
  bottom=4pt
]
\begin{Verbatim}
You are evaluating user-viewpoint understanding in a therapy session. For each checkpoint, identify the current speaker's viewpoint about another on-screen person in this turn. Ask directly: who is expressing the view, who is the view about, and what is that view. The source should normally be the row's primary_speaker, and if there are multiple distinct viewpoints toward different people in the same row, output all of them. Do not use future rows or outside context. Return short, concrete English viewpoint phrases such as "she is perpetually dissatisfied and impossible to please," along with the corresponding source-target pairs and a short reason.
\end{Verbatim}
\end{tcolorbox}

\clearpage
\twocolumn

\end{document}